\pdfoutput=1
\documentclass[11pt, letterpaper, logo, onecolumn, copyright, numbering, ctexart]{paper_improved}

\usepackage{soul}
\usepackage{xcolor}

\usepackage{booktabs}
\usepackage{enumitem}
\usepackage{array}
\usepackage{colortbl}
\usepackage{xcolor}
\usepackage{caption}
\usepackage{tabularray}
\usepackage[utf8]{inputenc} % allow utf-8 input
\usepackage[T1]{fontenc}    % use 8-bit T1 fonts
\usepackage{hyperref}       % hyperlinks
\usepackage{url}            % simple URL typesetting
\usepackage{booktabs}       % professional-quality tables
\usepackage{amsfonts}       % blackboard math symbols
\usepackage{nicefrac}       % compact symbols for 1/2, etc.
\usepackage{microtype}      % microtypography
\usepackage{xcolor}         % colors
\usepackage{graphicx}

\usepackage{natbib}
\usepackage{verbatim}
\usepackage{inconsolata}
\usepackage{tabularx}
\usepackage{float}
\usepackage{algorithmic}
\usepackage{multirow}
\usepackage{listings}
\usepackage{xcolor}
\usepackage{subfigure}
\usepackage{subcaption}
\usepackage{amsmath}
\usepackage{amssymb}
\usepackage{array}
\usepackage{CJKutf8}
\usepackage[ruled,linesnumbered]{algorithm2e}
\usepackage[T1]{fontenc}
\definecolor{lightgray}{rgb}{0.95, 0.95, 0.95}
\definecolor{darkgray}{rgb}{0.4, 0.4, 0.4}
\definecolor{backcolour}{rgb}{0.95,0.95,0.92}
\definecolor{myblue}{rgb}{0.2, 0.4, 0.8} % Example blue color
\definecolor{mygreen}{rgb}{0.2, 0.6, 0.2} % Example green color

\usepackage{amsmath,amssymb,amsfonts}

\let\cite\citep

\title{TransBench: Benchmarking Machine Translation for Industrial-Scale Applications}

\reportnumber{} % Leave blank if n/a

\author{Haijun Li\textsuperscript{1*}, Tianqi Shi\textsuperscript{1*}, Zifu Shang\textsuperscript{1}, Yuxuan Han\textsuperscript{1}, Xueyu Zhao\textsuperscript{1}, Hao Wang\textsuperscript{1}, Yu Qian\textsuperscript{1}, Zhiqiang Qian\textsuperscript{1}, Linlong Xu\textsuperscript{1}, Minghao Wu\textsuperscript{1}, Chenyang Lyu\textsuperscript{1}, Longyue Wang\textsuperscript{1}, Gongbo Tang\textsuperscript{2}, Weihua Luo\textsuperscript{1}, Zhao Xu\textsuperscript{1}, Kaifu Zhang\textsuperscript{1}\\
\textsuperscript{1} Alibaba International Digital Commerce\\
\textsuperscript{2} Beijing Language and Culture University
}

\footnotetext[1]{Equal Contribution.}
\begin{abstract}

Machine translation (MT) has become indispensable for cross-border communication in globalized industries like e-commerce, finance, and legal services, with recent advancements in large language models (LLMs) significantly enhancing translation quality. However, applying general-purpose MT models to industrial scenarios reveals critical limitations due to domain-specific terminology, cultural nuances, and stylistic conventions absent in generic benchmarks. Existing evaluation frameworks inadequately assess performance in specialized contexts, creating a gap between academic benchmarks and real-world efficacy. To address this, we propose a three-level translation capability framework: (1) Basic Linguistic Competence, (2) Domain-Specific Proficiency, and (3) Cultural Adaptation, emphasizing the need for holistic evaluation across these dimensions. We introduce TransBench, a benchmark tailored for industrial MT, initially targeting international e-commerce with 17,000 professionally translated sentences spanning 4 main scenarios and 33 language pairs. TransBench integrates traditional metrics (BLEU, TER) with Marco-MOS, a domain-specific evaluation model, and provides guidelines for reproducible benchmark construction. Our contributions include: (1) a structured framework for industrial MT evaluation, (2) the first publicly available benchmark for e-commerce translation, (3) novel metrics probing multi-level translation quality, and (4) open-sourced evaluation tools. This work bridges the evaluation gap, enabling researchers and practitioners to systematically assess and enhance MT systems for industry-specific needs.
\end{abstract}

\begin{document}

\maketitle

\vspace{2\baselineskip} 

\begin{figure}[h]
    \centering
    \includegraphics[width=0.9\linewidth]{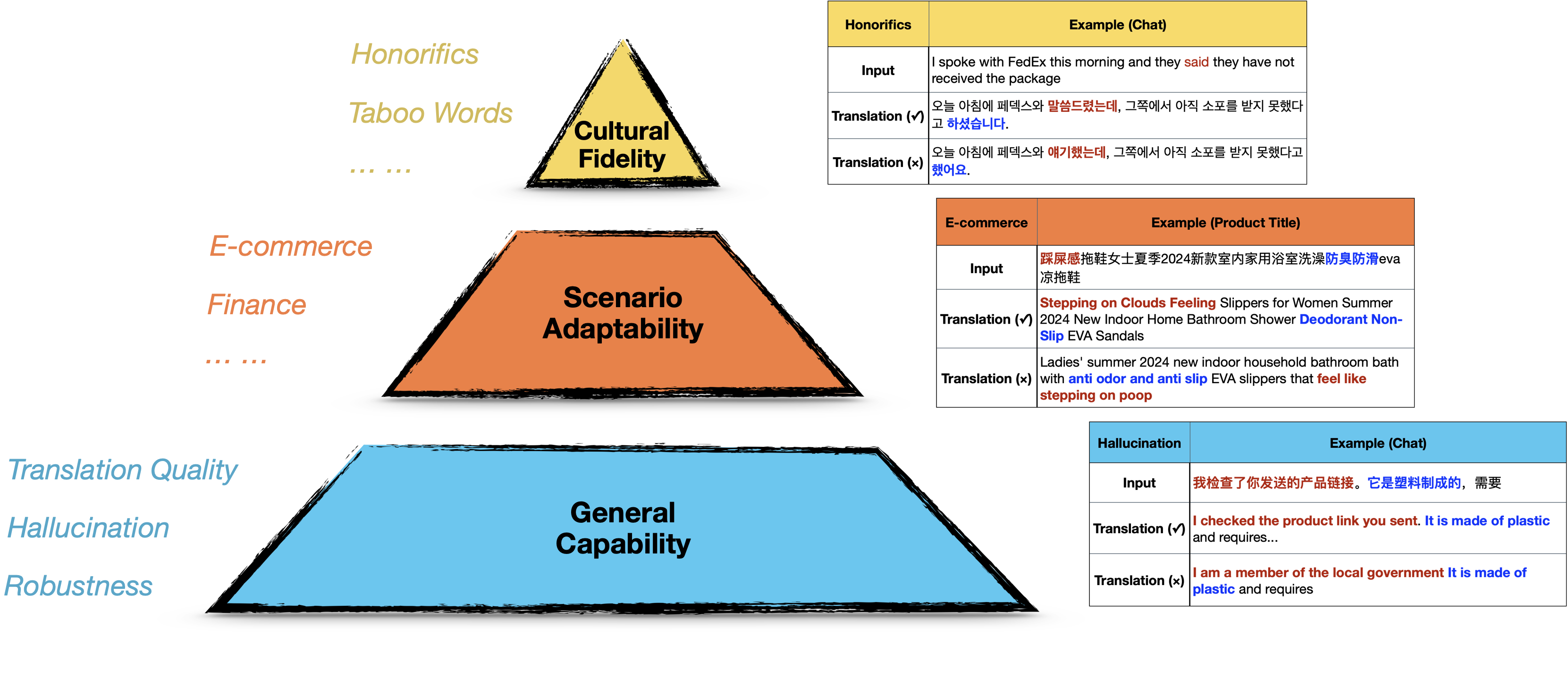}
    \caption{The Overall Framework of TransBench Benchmark and Evaluate Matrices. The Benchmark Covers Multilingual Sentences and the Evaluation Matrices are Fair and Comprehensive.}
    \label{fig:vs}
\end{figure}

% \cleardoublepage
% \phantomsection
% \tableofcontents

\newpage
\section{Introduction}
\label{sec:intro}
% background
In the era of rapid globalization and the booming digital economy, machine translation (MT) has emerged as a pivotal technology for enabling cross-border communication on an unprecedented scale. It is crucial in service industries such as international E-commerce, finance, and legal services, where high-quality MT is fundamental to connecting global users, expanding international markets, and ensuring the accurate and effective delivery of information across language barriers. Significant advancements have been made in MT research in recent years, particularly with the advent of large language models (LLMs)~\cite{achiam2023gpt,hurst2024gpt4o,guo2025deepseek,liu2024deepseek,grattafiori2024llama} (such as GPT-4~\cite{achiam2023gpt,hurst2024gpt4o} and DeepSeek~\cite{liu2024deepseek,guo2025deepseek}), leading to remarkable improvement of translation quality, especially in terms of contextual coherence. 

% challenges in industrial MT
However, applying general-purpose MT models directly to real-world industrial application scenarios often reveals significant limitations. 
This is primarily because texts in these industrial contexts exhibit high domain specificity, involving technical jargon, fixed expressions, particular styles, and formats not commonly found in general text (e.g., news articles). Furthermore, these texts are deeply embedded within specific cultural contexts, requiring more than just linguistic transfer but also nuanced adaptation to resonate with the target audience. 
Existing models, predominantly trained and evaluated on generic data, frequently struggle to accurately handle specialized terminology, maintain appropriate domain-specific conventions, or effectively considering cross-cultural information. 

% evaluation gap
Crucially, the prevailing MT benchmark datasets and evaluation methodologies, do not adequately capture the complexity and specific requirements of industrial translation tasks. 
The WMT (Conference on Machine Translation)~\citep{wmt-2024-1} series and IWSLT~\citep{iwslt-1-2024} series are the most prominent and highly influential annual benchmarks for evaluating general machine translation systems. These benchmarks often lack realistic, domain- and culturally-representative data. 
Furthermore, automatic evaluation metrics, such as BLEU~\citep{papineni-etal-2002-bleu}, METEOR~\citep{banerjee-lavie-2005-meteor}, TER~\citep{snover-etal-2006-study}, chrF~\cite{popovic2015chrf}, COMET~\cite{rei2020comet}, etc., struggle to effectively measure performance along critical dimensions such as the accuracy of domain-specific terminology, adherence to specific styles, and the successful handling of cross-cultural adaptation. 
Consequently, a significant evaluation gap exists between observed performance on academic benchmarks and real-world effectiveness in industrial applications, making it challenging for both researchers and practitioners to accurately assess and improve MT quality for industry-specific use cases. 

% introduce the 3-level translation framework
To bridge this evaluation gap and provide a more structured understanding of the multifaceted nature of industrial translation, we propose a framework that views translation capability across three interconnected levels: (1) \textbf{Basic Linguistic Competence}, the ability to accurately map basic linguistic structures, grammar, and common vocabulary between languages in general contexts; (2) \textbf{Domain-Specific Proficiency}, the ability to understand and accurately translate texts adhering to specific domain conventions, terminology, and stylistic requirements (e.g., e-commerce product specifications, financial reports, legal contracts); and (3) \textbf{Culture Adaptation}, the ability to navigate subtle cultural nuances, adapt content for the target audience's cultural context. We posit that achieving robust, industry-ready MT requires models to possess high-level competence across all three of these interwoven dimensions. 

% introduce Transbench 
Motivated by this framework and the critical need for more relevant evaluation, we introduce \textbf{TransBench}, a novel benchmark specifically designed to evaluate machine translation models for real-world industrial application scenarios. 
TransBench is constructed based on our proposed three-level framework, aiming to probe model capabilities at each level. We initially focus on the highly representative and challenging domain of International E-commerce. We plan to expand TransBench to include other critical domains such as Finance and Law in future work. 
It comprises 17,000 professionally translated and verified sentences spanning 4 authentic E-commerce scenarios and 33 language pairs, and a comprehensive suite of evaluation metrics, including BLEU, METEOR, TER, chrF, Marco-MOS (an evaluation model tailored for E-commerce MT), etc. 

% main contributions
In this paper, we make the following key contributions:
\begin{itemize}[leftmargin=*,topsep=0.1em,itemsep=0.1em,parsep=0.1em]
    \item We propose and articulate a structured three-level framework for understanding and evaluating machine translation capability in industrial application scenarios. 
    \item We introduce TransBench, the first publicly available benchmark specifically designed based on our multi-level framework to evaluate MT models for industrial use cases, starting with the International E-commerce domain.
    \item We propose and implement a comprehensive set of evaluation metrics and methodologies designed to probe machine translation performance specifically across the three levels. 
    \item we introduce Marco-MOS, an automated translation quality assessment model, which benefits to the development of E-commerce MT. 
    \item we release the overall framework for benchmark construction, including translation quality assessment guidelines, algorithms, rules, manual annotation and proofreading processes, to help others quickly build benchmarks.
    \item we publicly release the evaluation framework used in this paper, to foster research in E-commerce machine translation and related. 
    % \item Furthermore, we demonstrate the effectiveness of Marco-LLM in any-to-any machine translation tasks, where Marco achieved superior performance for translating one low-resource language to another one.
\end{itemize}

\section{Related Work}
\label{sec:related_work}
% \paragraph{Large Language Models}

\paragraph{Machine Translation}

The development of machine translation (MT) has progressed through a series of influential approaches and technological innovations. Early MT efforts focused on rule-based machine translation (RBMT), which utilized linguistic rules and bilingual dictionaries to convert text between languages. However, these systems often struggled with linguistic ambiguities and exceptions \citep{knight-1993-building,hutchins-1993-latest,DBLP:conf/aaai/KnightL94,DBLP:journals/mt/LonsdaleMN94}. To address these challenges, statistical machine translation (SMT) emerged, leveraging large parallel corpora to learn probabilistic translation models based on word and phrase alignments \citep{DBLP:journals/csur/Lopez08,DBLP:books/daglib/0032677,DBLP:journals/llc/HearneW11}. SMT improved translation adequacy by statistically inferring likely translations, but it faced limitations in modeling long-range dependencies, producing fluent and natural text, and more. The advent of neural machine translation (NMT) introduced a paradigm shift by employing neural networks to model the translation process end-to-end \citep{DBLP:conf/nips/SutskeverVL14,cho-etal-2014-learning,DBLP:conf/icml/GehringAGYD17}. NMT systems, particularly those using encoder-decoder architectures with attention mechanisms \citep{DBLP:journals/corr/BahdanauCB14,DBLP:conf/nips/VaswaniSPUJGKP17}, significantly enhanced translation quality by capturing complex linguistic patterns and context. Noteworthy contributions in this area include MT systems such as Google Translate \citep{DBLP:journals/corr/WuSCLNMKCGMKSJL16,johnson-etal-2017-googles}, Facebook's M2M-100 \citep{DBLP:journals/jmlr/FanBSMEGBCWCGBL21}, NLLB \citep{DBLP:journals/corr/abs-2207-04672}, and SeamlessM4T \citep{DBLP:journals/corr/abs-2308-11596}. More recently, large language models (LLMs) that utilize transformer architectures and are pre-trained on extensive multilingual datasets have revolutionized the paradigm and set new benchmarks in MT by effectively handling context, disambiguation, and fluency across diverse languages \citep{wang-etal-2023-document-level,DBLP:conf/iclr/Xu0SA24,DBLP:journals/corr/abs-2402-17733,DBLP:conf/icml/XuSCTSDM024,DBLP:journals/corr/abs-2405-11804}, demonstrating the potential of LLMs to deliver high-quality translations and have expanded the horizons of cross-lingual understanding and communication.

\paragraph{MT Evaluation Benchmarks}
MT evaluation benchmarks serve as critical resources for assessing and comparing translation systems across diverse linguistic contexts. Traditional benchmarks like WMT \citep{wmt-2024-1}, IWSLT \citep{iwslt-1-2024}, and NIST \citep{przybocki2009nist} have established standardized multilingual test sets that enable consistent cross-system evaluation. These benchmark collections have evolved to address specific challenges in the field, including low-resource language pairs (e.g., FLORES \citep{goyal2022flores,nllb2024scaling}), domain-specific translation (e.g., TICO-19 for COVID-related content \citep{anastasopoulos2020tico}), and robustness to non-standard text (e.g., MTNT \citep{michel2018mtnt}). Contemporary benchmarks increasingly focus on challenging phenomena such as discourse-level coherence \citep{muller-etal-2018-large,lopes-etal-2020-document}, cultural nuance preservation, and handling of linguistic ambiguities that frequently challenge translation systems. Challenge sets targeting specific linguistic phenomena have emerged to provide fine-grained analysis of model capabilities beyond general fluency and adequacy \citep{deutsch2025wmt24++}. The development of multilingual benchmarks covering hundreds of languages has been particularly valuable for evaluating the cross-lingual transfer capabilities of machine translation systems, especially for language pairs with limited parallel data \citep{caswell2025smol}. Despite these advancements, significant gaps remain in benchmark coverage, particularly regarding dialectal variations, code-switching phenomena, and specialized domains where human expertise is required for accurate assessment.

\paragraph{MT Evaluation Metrics}
MT evaluation metrics provide frameworks for measuring translation quality, broadly categorized into automatic and human evaluation methods. Automatic metrics have evolved from lexical comparisons like BLEU \citep{papineni-etal-2002-bleu}, METEOR \citep{banerjee-lavie-2005-meteor}, and TER \citep{snover-etal-2006-study} to embedding-based approaches capturing semantic similarity. Neural metrics such as BERTScore \citep{zhang2019bertscore}, COMET \citep{rei2020comet}, and BLEURT \citep{sellam-etal-2020-bleurt} leverage pre-trained language models to better align with human judgments across languages. These metrics address limitations in capturing semantic equivalence, fluency, and adequacy without relying on exact matches. Recent advancements include reference-free metrics like COMET-QE \citep{rei-etal-2021-references}, document-level metrics for discourse coherence \citep{vernikos-etal-2022-embarrassingly,jiang-etal-2022-blonde}, and linguistically informed approaches for interpretability. Human evaluation remains the gold standard despite its cost, using protocols like direct assessment \citep{graham-etal-2013-continuous} and Multidimensional Quality Metrics (MQM) \citep{burchardt-2013-multidimensional} to identify translation issues. Large language models (LLMs) have introduced a paradigm shift, enabling evaluators like GEMBA \citep{kocmi-federmann-2023-large,kocmi-federmann-2023-gemba}, xCOMET \citep{colombo2023xcomet}, and MetricX \citep{juraska-etal-2024-metricx} to assess translations based on semantic equivalence, fluency, and appropriateness through prompting.

% \paragraph{Ours}
\section{A Three-Level Framework for Industrial MT Capability}
\label{sec:framework}

Building upon the challenges identified in industrial MT and the limitations of existing evaluation paradigms discussed in Section~\ref{sec:intro}, we propose a structured framework for understanding and assessing the multifaceted capabilities required for robust MT performance in real-world industrial application scenarios. This framework posits that effective industrial translation goes beyond mere linguistic transfer, necessitating proficiency across distinct yet interconnected levels. We delineate translation capability into three core dimensions: 1) Basic Linguistic Competence, 2) Domain-Specific Proficiency, and 3) Culture Adaptation. This layered perspective is informed by established translation theories and reflects the practical complexities encountered by both human translators and MT systems in professional settings. The overall corresponding relationship between evaluation data and the matrices are shown in the Table\ref{tab:transbench_data}. 

\subsection{Linguistic Competence} 

The foundational level of translation capability is Basic Linguistic Competence. This refers to the fundamental ability to accurately process the source language and generate grammatically correct and fluent text in the target language, adhering to basic linguistic rules and accurately mapping common vocabulary and simple sentence structures. This level is primarily concerned with the linguistic well-formedness and semantic fidelity at the word and sentence levels in a general context, independent of specialized domains or deep cultural nuances.

This level aligns with early linguistic theories of translation, which largely focused on the formal transfer between language systems \citep{catford1965linguistic}. It represents the core competence in bilingual text processing and is the target of evaluation for basic linguistic quality metrics such as grammaticality, fluency, and literal accuracy of common language. 

Basic Linguistic Competence is the essential prerequisite for any usable translation. Errors at this level, such as grammatical mistakes, awkward phrasing, or incorrect translation of common words, immediately degrade translation quality, undermine credibility, and can render the text incomprehensible or unprofessional, regardless of its domain or cultural context. Ensuring a solid foundation at this level is non-negotiable for industrial applications. 

\subsection{Domain-Specific Proficiency}

The second level is Domain-Specific Proficiency. This involves the ability to accurately understand and translate texts that pertain to a specific industry or subject matter. It goes beyond general linguistic skills to encompass a deep understanding and correct application of: 
1) Specialized Terminology and Jargon: Accurately translating terms unique to a domain (e.g., product specifications, financial instruments, legal clauses). 
2) Domain-Specific Conventions and Style: Adhering to the typical phrasing, sentence structures, and stylistic norms of a particular field (e.g., the formal, precise language of legal texts vs. the persuasive tone of marketing copy). 
3) Knowledge of Domain Concepts: Understanding the underlying concepts to ensure accurate and contextually appropriate translation, even when direct linguistic equivalents are not available. 

This level is strongly supported by Functionalist translation theories, particularly Skopos Theory \citep{reiss1984grundlegung}, which emphasizes that the translation strategy is determined by the purpose of the translation in its target context. Domain-specific texts have specific functions (e.g., inform, instruct, persuade within a field), and achieving this function requires mastering domain conventions. It also relates to the linguistic concept of Register, which describes how language varies according to situation, topic, and relationship \citep{hatim1990discourse}. Research in Language for Specific Purposes (LSP) provides further theoretical grounding for the unique linguistic features of different domains. 

Domain-Specific Proficiency is critical for accuracy, clarity, and credibility in professional industrial settings. Incorrect domain terminology or inappropriate style can lead to serious misunderstandings, legal issues, financial losses, or failure to effectively convey essential product information in areas like e-commerce. Accurate domain handling is a hallmark of professional-quality translation. 

\subsection{Culture Adaptation}

The third level is Culture Adaptation. This encompasses the ability to understand and appropriately handle the cultural dimensions embedded in the source text and required for the target context. This goes beyond linguistic and domain-specific accuracy to consider how the text will be perceived and function within the target culture. Key aspects include:
1) Understanding Cultural Nuances: Recognizing implicit meanings, allusions, idioms, and humor tied to the source culture.
2) Adapting Cultural References: Deciding how to handle culturally specific items (e.g., customs, values, historical events, currency, units) to ensure they are understandable and acceptable to the target audience. 
3) Adjusting Communication Style: Adapting tone, politeness levels, and persuasive strategies to align with target cultural norms. 
4) Ensuring Functional Appropriateness: Ultimately, ensuring the translation effectively achieves its intended function (the Skopos) within the constraints and expectations of the target culture. This often involves elements of localization, where content is tailored to a specific locale. 

This level is fundamentally aligned with the Cultural Turn in translation studies \citep{bassnett1990translation}, which emphasizes the role of translation as a cross-cultural mediation process influenced by power, ideology, and cultural norms. It heavily draws on Functionalist approaches (Skopos Theory, Functional Equivalence \citep{nida1964toward}), as achieving the desired function often necessitates cultural adaptation to resonate with the target audience and context. 

Culture Adaptation is crucial for the efficacy and acceptance of translations in international markets, particularly in direct-to-consumer applications like international e-commerce marketing. Failure to adapt culturally can result in misunderstandings, unintended offense, brand damage, or simply render the translation ineffective or irrelevant to the target audience. This level ensures the translation is not only correct but also appropriate, engaging, and persuasive within its new cultural home.

\subsection{Interconnections and Framework Utility}

These three levels of translation capability are hierarchical but interconnected. Basic Linguistic Competence serves as the essential foundation for both Domain-Specific Proficiency and Culture Adaptation; errors at lower levels will invariably undermine performance at higher levels. Domain-Specific Proficiency relies on strong linguistic skills but applies them within a specialized context, often using language in ways dictated by domain conventions. Culture Adaptation operates across lower levels' outputs, ensuring that even technically accurate, domain-appropriate text is culturally suitable and functionally effective. A truly high-quality industrial translation requires a delicate balance and synergy across all three levels. 

This framework offers a structured, theoretically-grounded view of translation capability that aligns with the evolution of translation theories from purely linguistic to functional and cultural perspectives. It provides a systematic way to categorize translation challenges and errors. 
For industrial applications, this framework allows for the deconstruction of complex translation quality issues into more manageable components. It helps identify specific weaknesses in MT models. This targeted diagnosis is crucial for guiding model improvement efforts, designing more specific training data, and ensuring evaluation effectively probes the capabilities most critical for industrial success. 

In line with this framework's principles, the TransBench benchmark is specifically designed to provide data and evaluation methodologies that allow for the assessment of machine translation models' performance at each of these three crucial levels, as detailed in the following sections.

\section{TransBench Data}

\label{sec:transbench_data}

% In this chapter, we will introduce Transbench in detail, including the overall design architecture~\ref{sec:framework}, introduction to data sources~\ref{sec:data_source}, the overall annotation process (including algorithms, rule cleaning, manual annotation, special processing)~\ref{sec:data_processing}, and data statistics of the dataset~\ref{sec:data_statistics_and_analysis}.

Following the three-level framework for industrial MT capability introduced in Section~\ref{sec:framework}, we have constructed TransBench, a comprehensive benchmark featuring multiple evaluation datasets. 

\subsection{Dataset Structure}

The TransBench dataset is organized into distinct sections corresponding to key aspects of industrial MT capability, allowing for targeted evaluation based on our proposed framework. Table~\ref{tab:transbench} demonstrates the details of the TransBench. The dataset is structured to support evaluation of:

\textbf{Scenario Adaptability}: To evaluate Scenario Adaptability, which directly corresponds to Domain-Specific Proficiency, we include datasets specifically curated from distinct industrial sectors where specialized language is paramount. Currently, this category provides data for assessing translation quality in the International E-commerce domain and the Financial domain. These datasets allow for a focused assessment of how well models handle domain-specific terminology, style, and conventions.

\textbf{General Capability}: This category includes datasets designed to assess foundational aspects of model reliability and common pitfalls that can occur regardless of specific domain. This involves probing fundamental translation quality on general texts, as well as evaluating hallucination (generating content not present in the source) and overall robustness to variations or noise in the input. Ensuring reliability at this level is fundamental for building trust in industrial MT systems.

\textbf{Cultural Fidelity}: Finally, the Cultural Fidelity category comprises datasets constructed to probe model performance related to Culture Adaptation. These datasets focus on challenging socio-pragmatic and cultural phenomena that require sensitive handling to ensure appropriateness and effectiveness in the target culture, evaluating the correct translation and usage of Honorifics in languages where they are crucial, and assessing the appropriate handling of potentially sensitive or taboo words in different target cultures.

This multi-faceted dataset structure, grounded in our three-level framework, allows for a granular and insightful evaluation of MT models, enabling researchers and practitioners to pinpoint strengths and weaknesses across the different layers of capability essential for successful industrial deployment. The following subsections detail the specific characteristics and construction methodology for each of these dataset categories.

% \usepackage{multirow}
% \usepackage[normalem]{ulem}
\begin{table}[]
\centering
\begin{tabular}{c|c|c|c}
\hline
\rowcolor[HTML]{EFEFEF} 
\textbf{Evaluation Structure} & \textbf{Subdimension} & \textbf{\# Datasets} & \textbf{Evaluation Metrices} \\ \hline
 & \begin{tabular}[c]{@{}c@{}}E-commerce Industry \\ Translation Quality Assessment\end{tabular} & 17k & \begin{tabular}[c]{@{}c@{}}E-MOS \&\\ Common QE Metrices\end{tabular} \\ \cline{2-4} 
\multirow{-2}{*}{Scenario Adaptability} & \begin{tabular}[c]{@{}c@{}}Finance Industry Translation \\ Quality Assessment\end{tabular} & 12k & \begin{tabular}[c]{@{}c@{}}F-MOS \&\\ Common QE Metrices\end{tabular} \\ \hline
 & \begin{tabular}[c]{@{}c@{}}Translation Robustness \\ Assessment\end{tabular} & 2.6k & BLEU \\ \cline{2-4} 
\multirow{-2}{*}{General Capability} & \begin{tabular}[c]{@{}c@{}}Translation Hallucination\\ Rate Assessment\end{tabular} & 29k & Hallucination Rate \\ \hline
 & \begin{tabular}[c]{@{}c@{}}Taboo Word \\ Translation Assessment\end{tabular} & 232 & Accuracy \\ \cline{2-4} 
\multirow{-2}{*}{Cultural Fidelity} & \begin{tabular}[c]{@{}c@{}}Respectful Language \\ Translation Assessment\end{tabular} & 107 & Accuracy \\ \hline

\end{tabular}
\caption{The overall correspondence between data and evaluation methods, which also includes detailed information about the data and evaluation plan, such as data quantity and evaluation matrix.}
\label{tab:transbench_data}
\end{table}

\subsection{Scenario Adaptability Dataset}
% \subsubsection{Data Source}
\subsubsection{Comprehensive Multilingual Industrial Scenarios Data Source}

\begin{figure}[h]
    \centering
    \includegraphics[width=0.6\linewidth]{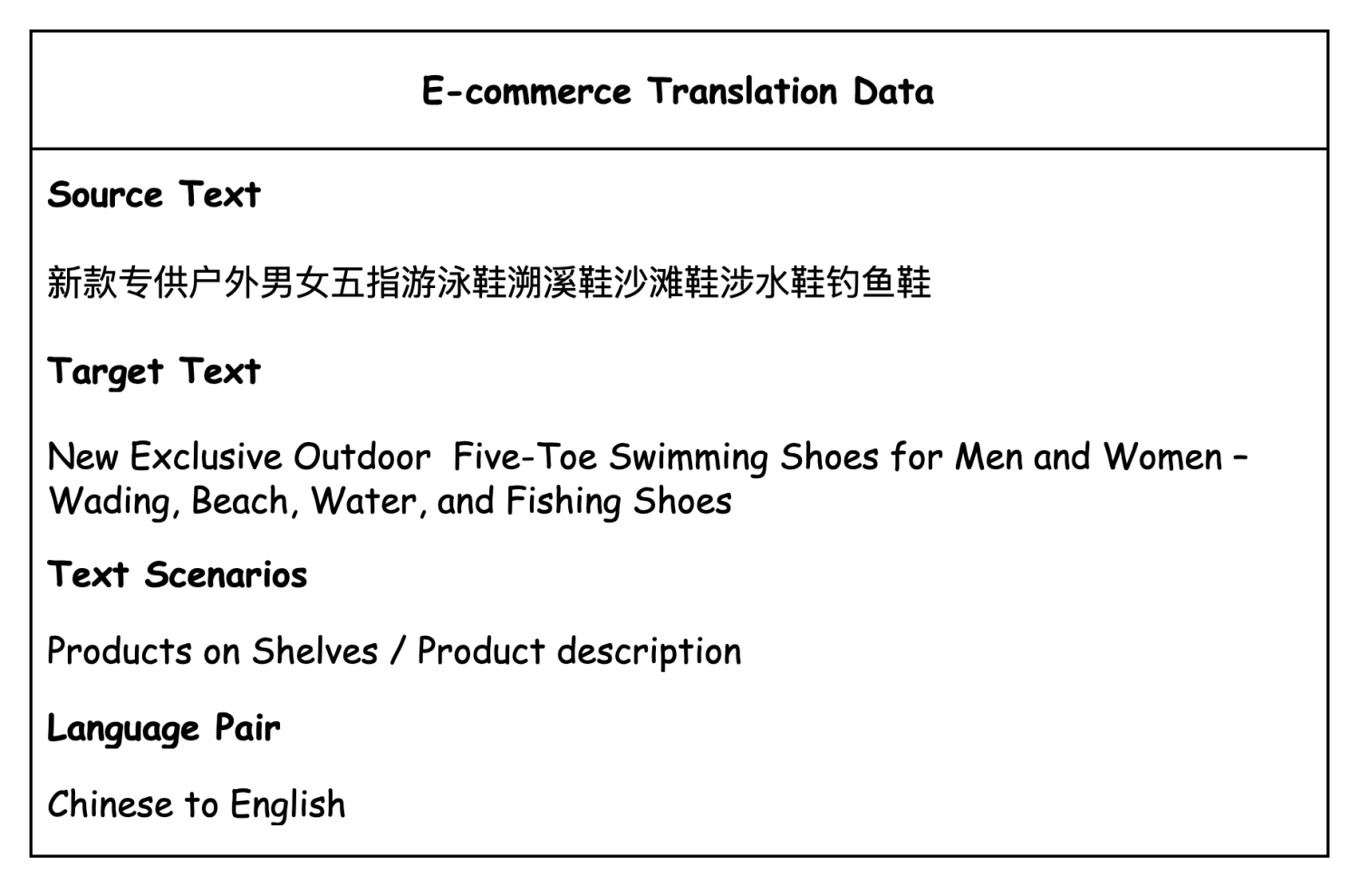}
    \caption{Data sample in the e-commerce field, which translated from Chinese to English. The text is sampled from the product listing-product description category. The source text represents the original input text, and the target text represents the text translated by the language expert.}
    \label{fig:sample1}
\end{figure}

\begin{figure}[h]
    \centering
    \includegraphics[width=0.6\linewidth]{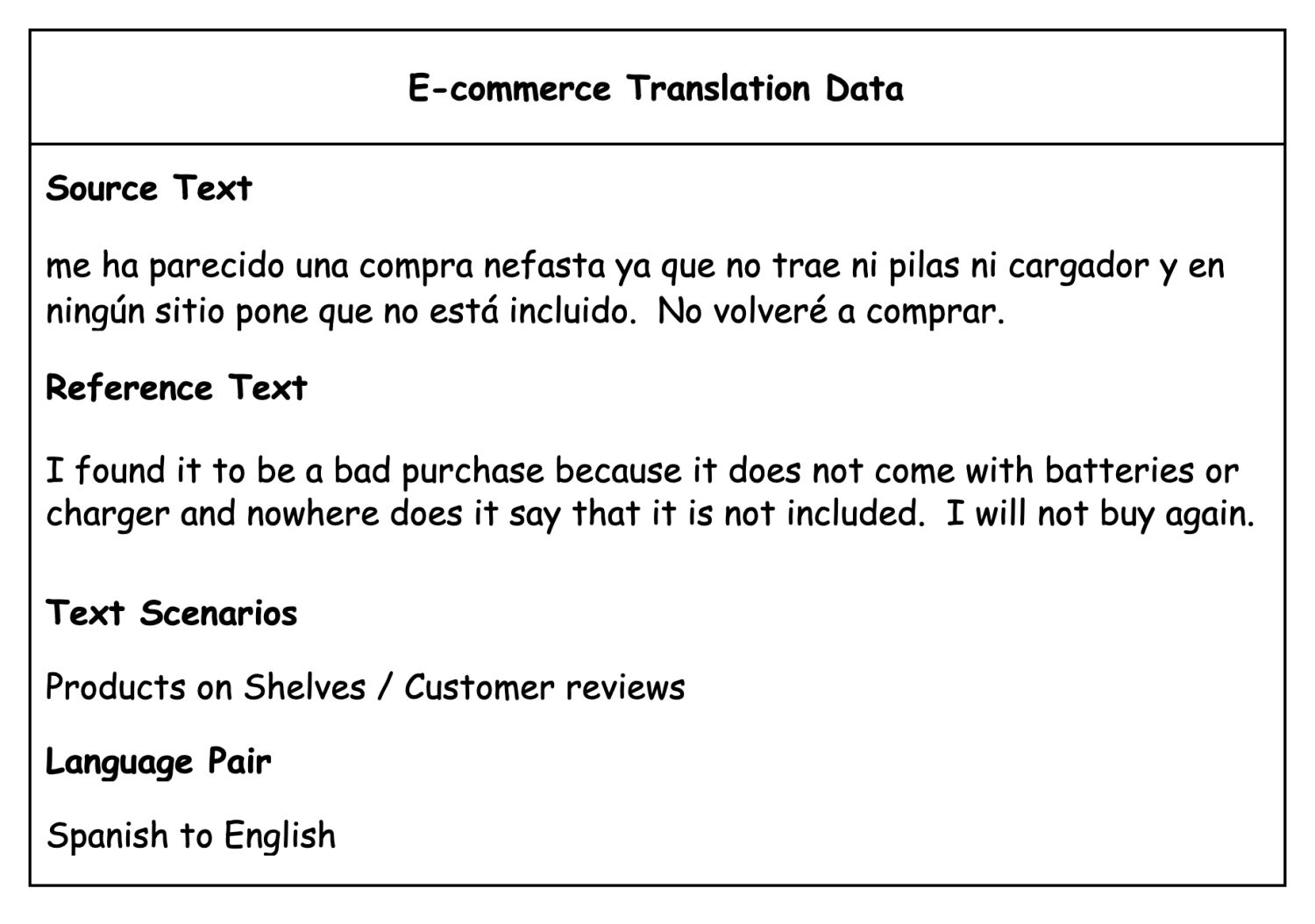}
    \caption{Data sample in the e-commerce field, which translated from Spanish to English. The text is sampled from the product customer reviews category. The source text represents the original input text, and the target text represents the text translated by the language expert.}
    \label{fig:sample2}
\end{figure}

\paragraph{E-Commerce Scenarios}
% \noindent
The E-commerce dataset contains 17k translation data, which are extracted from real data of four major E-commerce scenarios and eight sub-scenarios and obtained through manual annotation. A total of 33 language pairs are involved.
E-commerce scenarios throughout the entire chain of E-commerce
In terms of the overall chain of e-commerce, we select from a large amount of real e-commerce data, and select representative high-quality data after multiple rounds of intelligent processing. The data samples are shown in Figure \ref{fig:sample1} and Figure\ref{fig:sample2} separately.
\begin{itemize}[leftmargin=*,topsep=0.1em,itemsep=0.1em,parsep=0.1em]
    \item \textbf{Products on Shelves:} Mainly includes the description, common questions, and e-commerce query.
    \item \textbf{Marketing Promotion:} Includes two sub-scenarios: search engine advertising and push messaging.
    \item \textbf{Customer Service:} Mainly includes sub-scenarios such as customer service QA.
    \item \textbf{Evaluation and Dissemination:} Mainly includes two sub-scenarios: user comments and replies.
    % \item Furthermore, we demonstrate the effectiveness of Marco-LLM in any-to-any machine translation tasks, where Marco achieved superior performance for translating one low-resource language to another one.
\end{itemize}

\paragraph{Finance Scenarios}
From four major financial scenarios, it is expected that there will be 60 language directions in total, 200 data in each language direction, and the overall recovery of translation and annotation data sets is expected to be 12k. The four major scenarios include: \textbf{financial natural language processing, financial scenario computing, financial analysis and interpretation, and financial compliance and security.}

\paragraph{Multiligual Dimension}
In terms of language coverage, TransBench covers a total of 16 languages, including: Chinese, English, German, French, Spanish, Russian, Japanese, Korean, Indonesian, Hindi, Thai, Portuguese, Turkish, Kazakh, Arabic, Malay. 

In terms of language pair coverage, TransBench covers 60 languages pairs totally, which include: zh-en, en-zh, de-en, en-de, de-zh, zh-de, fr-en, en-fr, fr-zh, zh-fr, es-en, en-es, es-zh, zh-es, es-pt, ru-en, en-ru, ru-zh, zh-ru, ja-en, en-ja, ja-zh, zh-ja, ko-en, en-ko, ko-zh, zh-ko, id-en, en-id, id-zh, zh-id, hd-en, en-hd, hd-zh, zh-hd, th- en, en-th, th-zh, zh-th, pt-en, en-pt, pt-zh, zh-pt, tr-en, en-tr, tr-zh, zh-tr,tr-ar, tr-de, kk-en, en-kk, kk-zh, zh-kk, ar-en, en-ar, ar-zh, zh-ar, ms-en, en-ms, ms-zh.

\subsubsection{Data Processing and Annotation}
In order to ensure the authority and high-quality of TransBench data, we generally adopt three steps: algorithm filtering, manual annotation, and special processing. The purpose of algorithm filtering is to remove duplicates and obviously low-quality data. Manual annotation is to use professional translators to directly annotate rather than use post-editing, even if it costs more. Special processing is for processing toxic, sensitive data.

\paragraph{Data process by machine}
\begin{itemize}[leftmargin=*,topsep=0.1em,itemsep=0.1em,parsep=0.1em]
    \item \textbf{Data Cleaning}. Given that the majority of our dataset is derived from E-commerce data from all scenarios in E-commerce, we implemented a comprehensive data processing strategy to ensure quality. Initially, we applied regular expression filtering to remove incorrect numbering in segmented outputs, HTML format outputs, emoji data, and hyperlinks or URL references.
    \item \textbf{Data Filtering}. To enhance the data quality, we employed a comprehensive pipeline based on seminal works for data quality filtering, effectively removing low-quality samples.
\end{itemize}

\paragraph{Mannual Annotation pipeline}
The translation annotation process in this study strictly follows a standard operating procedure to ensure the quality. 
\begin{itemize}[leftmargin=*,topsep=0.1em,itemsep=0.1em,parsep=0.1em]
    \item \textbf{Pre-Production Phase}. The translation guidelines mandate strictly manual execution, prohibiting machine translation and large language models (LLMs), while requiring literal phrase-level translation with preserved source text structures. Problematic source data must be systematically annotated, and unique textual characteristics retained to ensure high-quality, culturally appropriate, and diverse general-purpose datasets. 
    \item \textbf{Production Phase}.The translation production protocol enforces manual execution with phrase-level literal fidelity and source structure preservation, while implementing systematic annotation of non-conforming inputs and retention of stylistic artifacts. 
    \item \textbf{Post-Production Phase}. The quality assurance protocol employs double-blind cross-verification and expert bilingual evaluation through a structured scoring system (1-5 scale), complemented by dynamic terminology management and performance metrics, to ensure linguistic consistency, cultural compliance, and regulatory alignment in multilingual datasets. 
\end{itemize}

\paragraph{Specific process}
Considering of data protection, special desensitization is performed for personal information and sensitive information, as well as data detoxification and data desensitization.
\begin{itemize}[leftmargin=*,topsep=0.1em,itemsep=0.1em,parsep=0.1em]
    \item \textbf{Data desensitization:} Personal sensitive information and user data that appear in the data, such as user purchase information, merchant contact information, etc., are identified and desensitized.
    \item \textbf{Data detoxification:} Personal sensitive information and user data that appear in the data, such as user purchase information, merchant contact information, etc., are identified and desensitized.
    % \item \textbf{Special Considerations for E-commerce Translation:} Special Considerations for E-commerce Translation
    % \item \textbf{Special Considerations for Finance Translation:} Special Considerations for finance Translation
\end{itemize}

\subsection{General Capability Dataset}
In the General Capability Assessment dataset, we primarily examine two types of data. The first type is translation hallucination data, which includes five common issues related to translation hallucinations. The second type is translation robustness data, which consists of three levels of robustness testing: sentence-level, character-level, and word-level.

\subsubsection{Hallucination Dataset}
In industrial translation, the following common translation illusion problems may occur:
\begin{itemize}[leftmargin=*,topsep=0.1em,itemsep=0.1em,parsep=0.1em]
    \item \textbf{Repetitive Generation}. This phenomenon occurs when the translation model unnecessarily duplicates phrases or sentences in the output, leading to verbose and redundant translations. Repetitive generation can hinder readability and may indicate issues with the model’s ability to maintain coherence and control over the generated content.
    \item \textbf{Entire Sentence Untranslated or Mistranslated}. In some cases, the model fails to translate an entire sentence, either leaving it in the source language or producing a completely incorrect translation. This reflects a critical failure in the model’s understanding or processing of the input text, especially for complex or low-resource language pairs.
    \item \textbf{Output Language Mismatch}. This issue arises when the model generates text in a target language different from the one specified in the translation task. It indicates potential problems in language identification or control mechanisms within the model, particularly when handling multilingual translation systems.
    \item \textbf{Partial Translation Omission}. Partial translation omission refers to situations where certain segments or phrases within a sentence are left untranslated or ignored by the model. This issue may occur due to attention misalignment or difficulties in handling domain-specific or rare vocabulary items.
    \item \textbf{Under-translation or Over-translation}. Under-translation occurs when the translated output lacks information present in the source text, while over-translation involves adding extra, unintended content. Both phenomena reflect inaccuracies in the model’s comprehension and generation capabilities, impacting the fidelity and completeness of the translation.
\end{itemize}

\subsubsection{Robustness Dataset}
\begin{figure}[h]
    \centering
    \includegraphics[width=0.6\linewidth]{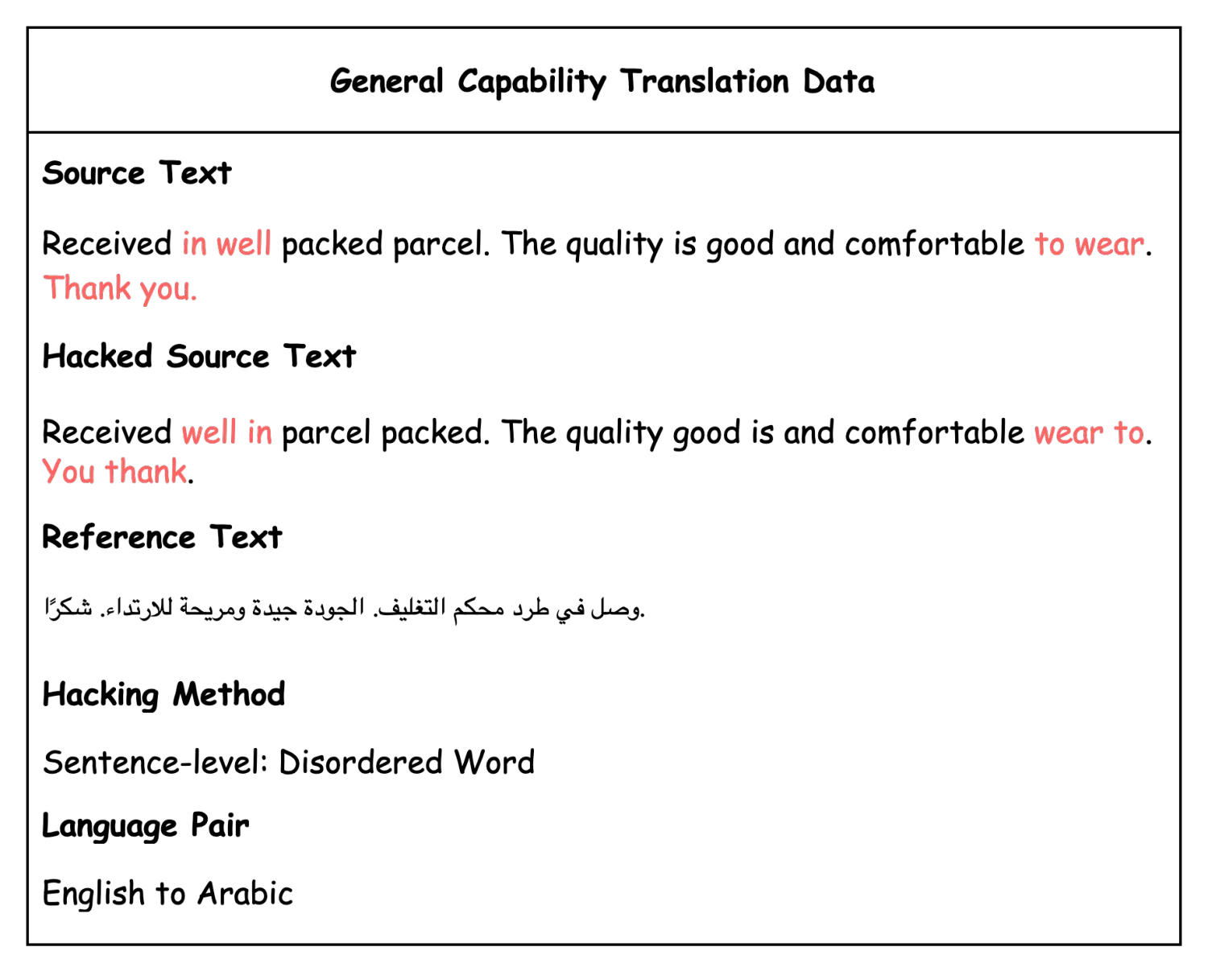}
    \caption{Sample data of General capality translation. This sample shows the data robust with the pattern of sentence-level-disordered word. It can be seen that the hacked source text scrambles some words from source text. It is hoped that the translation model can maintain the accuracy of the translation results.}
    \label{fig:sample3}
\end{figure}

\begin{figure}[h]
    \centering
    \includegraphics[width=0.6\linewidth]{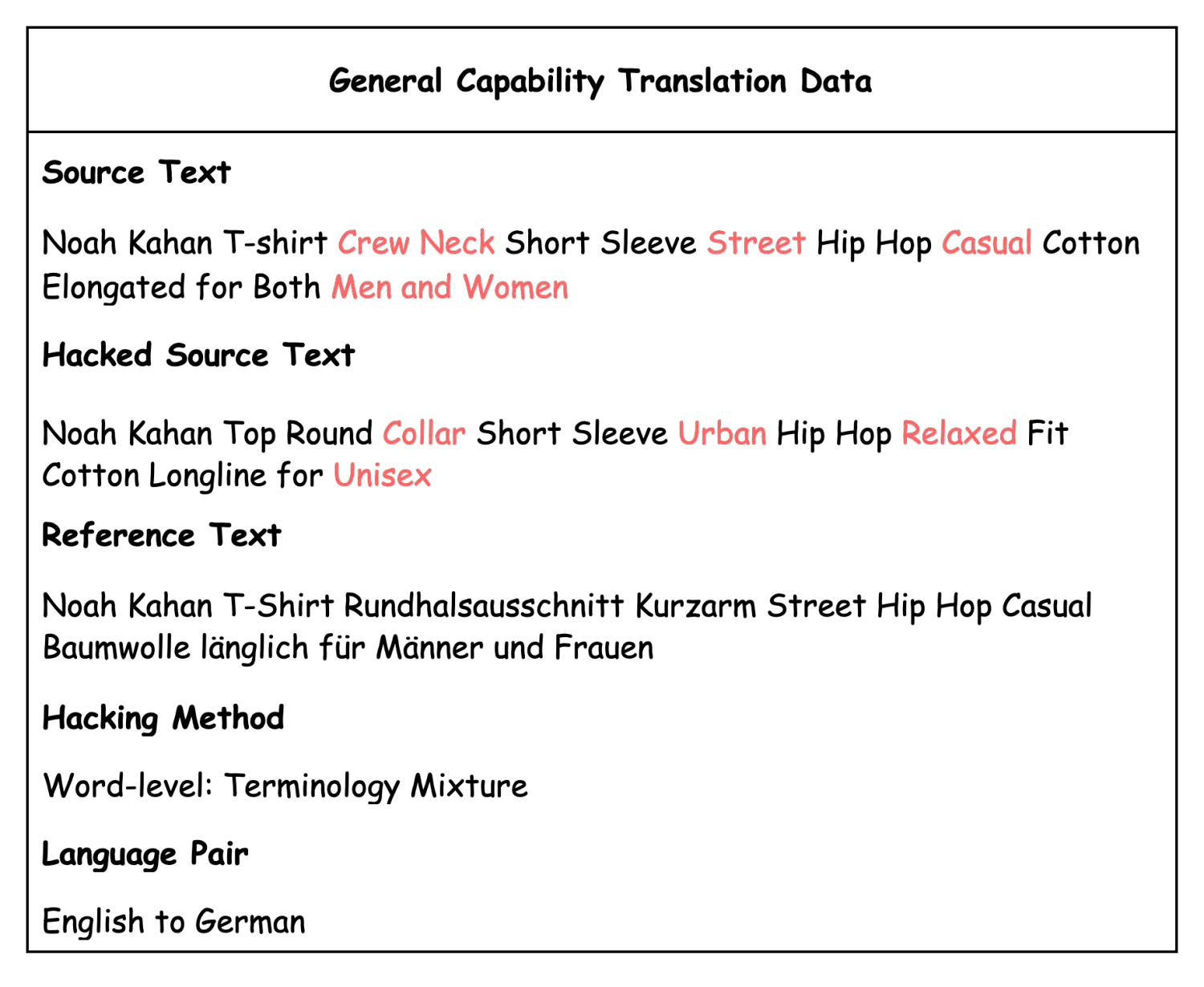}
    \caption{Sample data showing General capality translation. This sample shows the robust interference method of word-level terminology mixture. It can be seen that the Hacked Source Text has rewritten some terms in the Source Text at the semantic level.}
    \label{fig:sample4}
\end{figure}

\begin{figure}[h]
    \centering
    \includegraphics[width=0.6\linewidth]{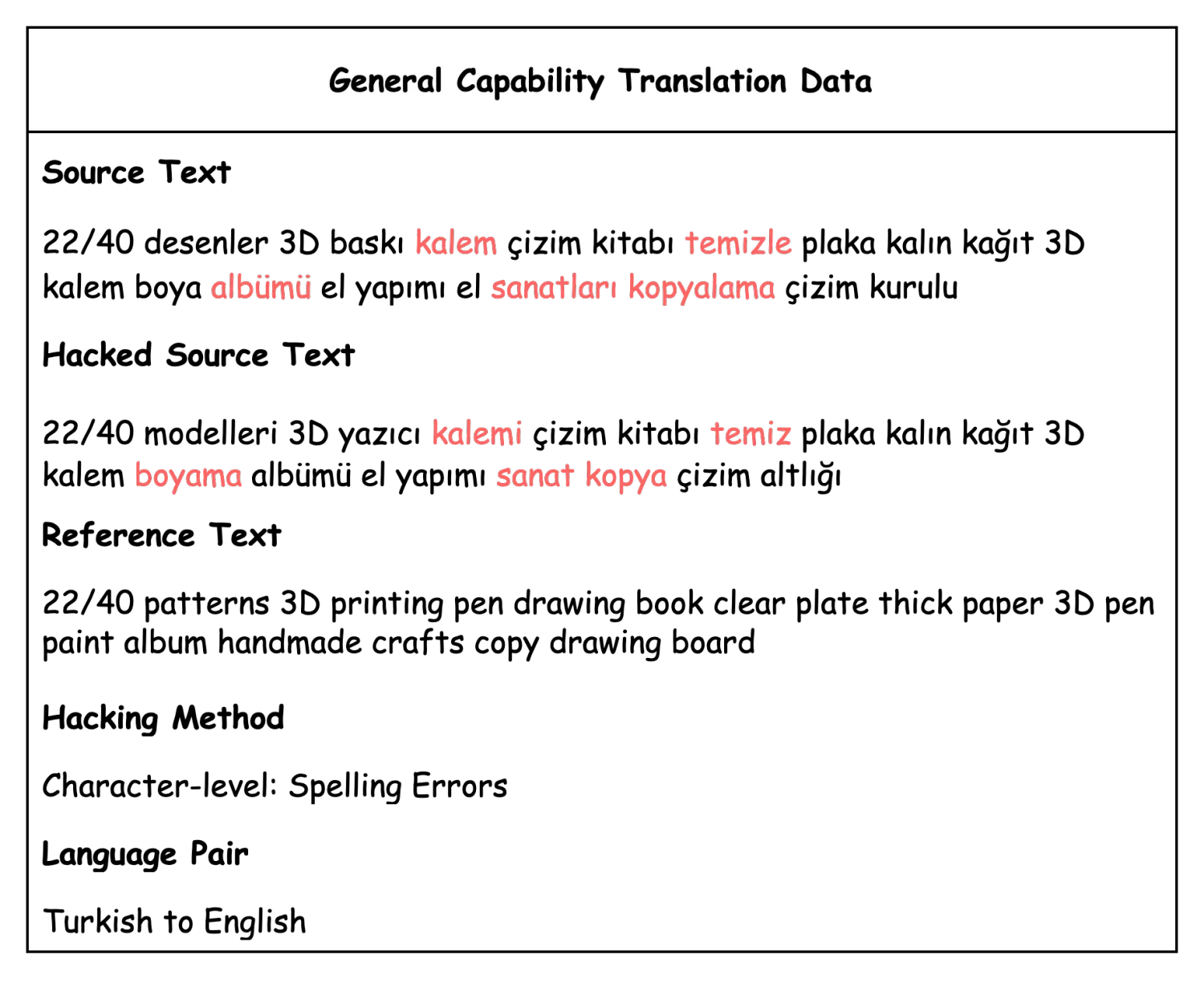}
    \caption{Sample data of General capality translation. This sample shows the data robust interference mode character-level: spelling errors. It can be seen that some words in the Hacked Source Text have random spelling errors.}
    \label{fig:sample5}
\end{figure}

In the actual working process of the translation model, the input source text does not always appear in a stable and normal state. Due to user habits, pasting errors, ambiguity of the original text and other problems, out of distribution source text often appears. At this time, if the translation model can stably output the correct translation results, it can prove the robustness of the translation system. The data samples are shown in Figure \ref{fig:sample3}, Figure\ref{fig:sample4} and Figure\ref{fig:sample5} separately.
\begin{itemize}[leftmargin=*,topsep=0.1em,itemsep=0.1em,parsep=0.1em]
    \item \textbf{Sentence-level: Disordered Word}. At the sentence level, translation models may produce syntactically ill-formed output due to incorrect word ordering. This issue often leads to a loss of coherence and readability, reflecting limitations in the model’s ability to capture and generate grammatically correct structures in the target language.
    \item \textbf{Character-level: Spelling Errors}. At the character level, models can generate text containing typographical or spelling mistakes, especially in languages with complex morphology or in the presence of rare or unseen character combinations. These errors may impair word recognition and overall translation quality.
    \item \textbf{Word-level: Terminology Mixture}. At the word level, inconsistent or incorrect use of terminology—particularly in domain-specific contexts—can be observed. Additionally, unintended code-mixing, where words from an unintended language appear in the output, may occur, undermining translation accuracy and domain adaptability.
\end{itemize}

% \subsubsection{Data Processing and Annotation}

\subsection{Cultural Fidelity Dataset}
The translation system involves a wide range of languages. In addition to maintaining the accuracy, robustness and stability of the translation information, it is also necessary to take into account the cultural attributes of different countries and regions to ensure that the translation results conform to the usage habits of the national culture of the translated language. In the cultural dimension evaluation data, we mainly examine the translation of taboo words and the translation of honorific systems.

\subsubsection{Honorifics Dataset}
\begin{figure}[h]
    \centering
    \includegraphics[width=0.6\linewidth]{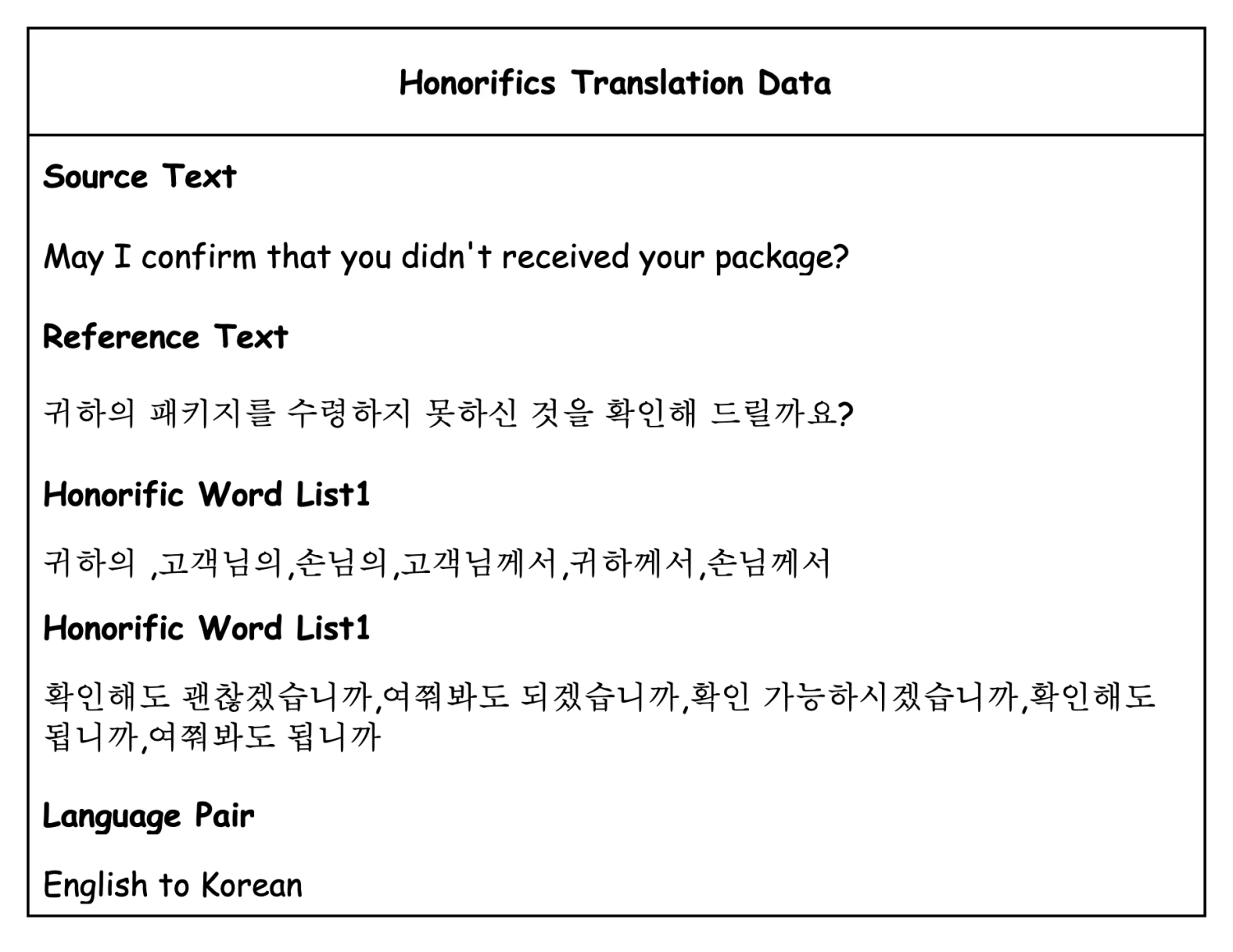}
    \caption{The display of honorific data in the Cultural Fidelity dataset. The source text is the text to be translated, and the reference text is the translated text using honorifics. Each data sample has multiple honorific units, and each unit must exist together in the translation result. Only one unit in each unit needs to exist in the translation result.}
    \label{fig:figure6}
\end{figure}

The honorific systems in both Korea and Japan emphasize politeness and respect, and are crucial components of their linguistic cultures. The data sample are shown in Figure \ref{fig:figure6}.

% \begin{CJK}{UTF8}{gbsn}
In Korea, honorifics are divided into grammatical and lexical categories. Grammatical honorifics involve adding specific suffixes or forms. Lexical honorifics include the use of special honorific vocabulary and verbs. Additionally, Korean honorifics are adjusted based on the listener's social status, age, and relationship.

In Japan, honorifics are mainly categorized into three types: Sonkeigo (respectful language), Kenjougo (humble language), and Teineigo (polite language). Sonkeigo is used to elevate the other person, Kenjougo is used to lower one's own status, and Teineigo is used for general polite expressions. Japanese honorifics are expressed through vocabulary and grammar, with common honorific verbs.
% \end{CJK}
The honorific systems of both countries are complex and nuanced, playing a significant role in everyday communication.

\subsubsection{Taboo Words Dataset}
\begin{figure}[h]
    \centering
    \includegraphics[width=0.6\linewidth]{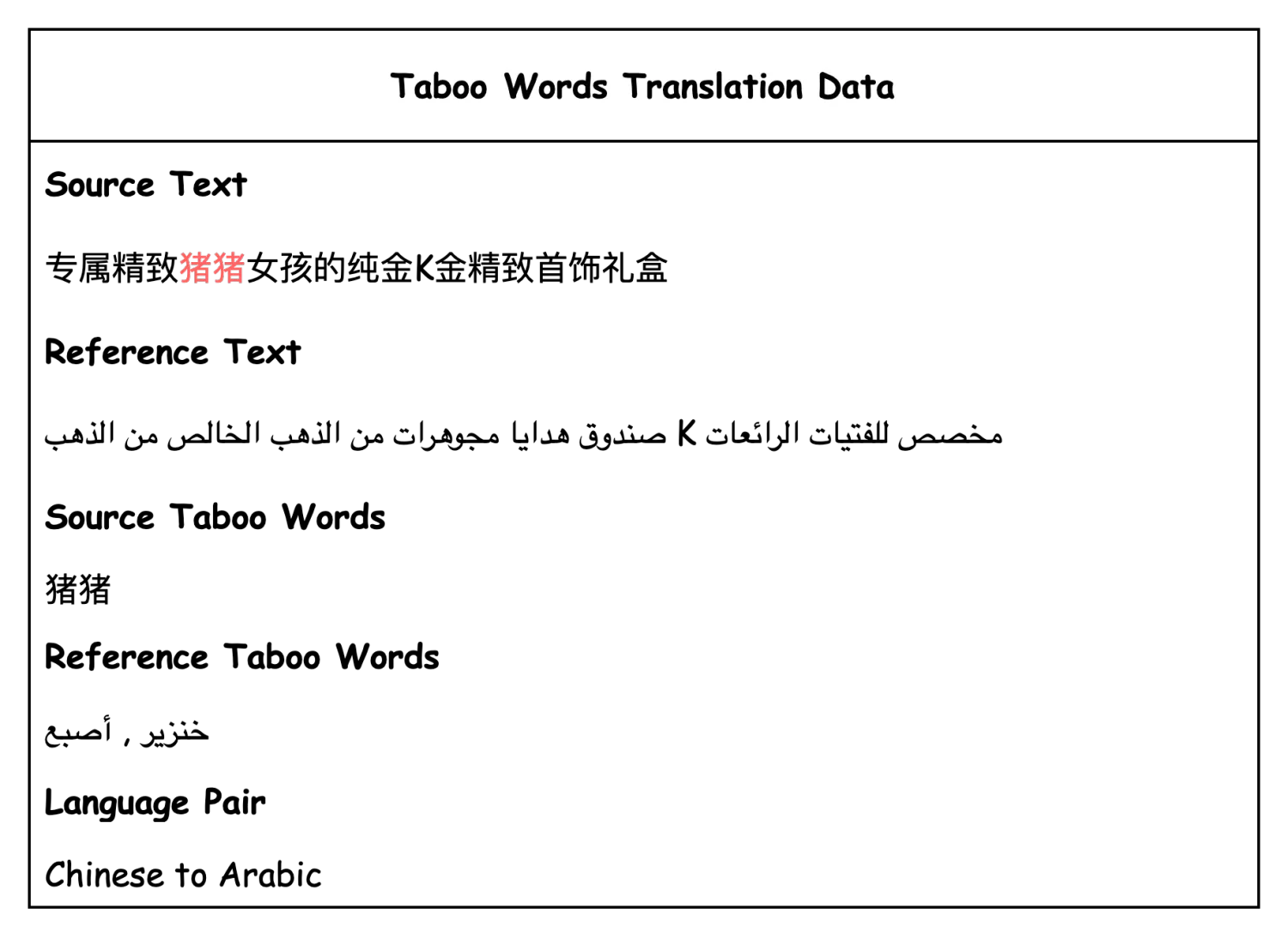}
    \caption{The Cultural Fidelity dataset shows the taboo words. The source text is the text to be translated, and the reference text is the translated text using honorifics. Each data sample has the reference taboo words before and after translation. The translation results should not contain reference taboo words.}
    \label{fig:figure7}
\end{figure}

Countries that use different languages often exhibit diverse customs and cultural practices. Due to variations in religious beliefs, political background, and social relationships, language usage may be subject to different taboos. For instance, in Arabic-speaking countries, where Islam is predominantly practiced, language pertaining to religious blasphemy, such as derogatory remarks about Allah or the Prophet Muhammad, is strictly prohibited. Similarly, in Indonesian and Malay-speaking regions, where the majority of the population is Muslim, religious taboos encompass various aspects of Islam, including disrespectful expressions towards Allah and religious rituals.The data sample are shown in Figure \ref{fig:figure7}.
% \subsubsection{Data Processing and Annotation}

\section{TransBench Evaluation Matrices}

\label{sec:transbench_evaluation_matrics}

% In this chapter, we will introduce Transbench in detail, including the overall design architecture~\ref{sec:framework}, introduction to data sources~\ref{sec:data_source}, the overall annotation process (including algorithms, rule cleaning, manual annotation, special processing)~\ref{sec:data_processing}, and data statistics of the dataset~\ref{sec:data_statistics_and_analysis}.

Evaluating the performance of MT models on the multi-faceted TransBench dataset requires a comprehensive suite of metrics capable of assessing the various dimensions of translation quality essential for industrial applications. Aligned with the structured dataset categories (Section~\ref{sec:transbench_data}), our evaluation methodology employs a combination of widely-used standard automatic metrics and novel, specialized approaches designed to probe model performance. 

\subsection{Overall Evaluation Matrices}

As illustrated in Table~\ref{tab:transbench_data}, our evaluation suite includes metrics tailored to each dataset category:

\textbf{General Capability Metrices}: We utilize a set of established automatic metrics to measure fundamental linguistic quality and reliability, which include popular N-gram based metrics such as BLEU and TER, character-based metrics like chrF, and model-based metrics like COMET-XXL which leverage large language models to assess translation quality. Additionally, specific metrics are employed for critical failure modes: Hallucination is quantified using a dedicated hallucination rate, and robustness is evaluated by measuring the degradation of baseline translation quality, typically assessed via a BLEU drop under perturbation. 

\textbf{Scenario Adaptability Metrices}: To provide a more relevant and accurate assessment of Scenario Adaptability, we introduce specialized automated evaluation models. We leverage domain-specific Quality Estimation (QE) models trained to predict human quality judgments within these contexts. Specifically, we utilize an translation quality estimation model called Marco-MOS, for the financial datasets. These models provide automated scores that are more sensitive to domain-specific correctness and appropriateness than general metrics. 

\textbf{Cultural Fidelity Metrices}: Evaluation for Cultural Fidelity focuses on the accurate and appropriate handling of specific challenging cultural phenomena. For datasets targeting aspects such as the correct translation and usage of Honorifics and the appropriate management of taboo words, performance is primarily measured through accuracy-based metrics. This involves evaluating whether the model correctly translates, renders, or handles these culturally sensitive items according to the target language and cultural norms.

Together, this comprehensive suite of multi-level evaluation metrics allows for a fine-grained diagnosis of MT model strengths and weaknesses across the different layers of capability essential for successful industrial deployment scenarios beyond general linguistic fluency. The following subsections detail the specific calculation and application of each metric category.

\subsection{General Capability Metrics}
\subsubsection{General Translation Quality Evaluation Matrices}
% \noindent
We selected common evaluation matrices commonly used in translation tasks, among which we focused on BLEU.

% \textbf{Comet~\cite{rei2020comet}}. 
% Comet(Crosslingual Optimized Metric for Evaluation of Translation) leverages multilingual pre-trained models (e.g., XLM-RoBERTa) to model the relationship between human ratings and vector space alignment via regression tasks. 

% % \noindent
% \textbf{Bleu~\cite{papineni2002bleu}}. 
% Bleu(Bilingual Evaluation Understudy) is a precision-oriented metric based on n-gram co-occurrence statistics, incorporating a brevity penalty (BP) to penalize overly short translations.

% % \noindent
% \textbf{chrF~\cite{popovic2015chrf}}. 
% chrF(Character n-gram F-score) is a character-level metric computing a weighted harmonic mean of precision (chrP) and recall (chrR) using n-grams, emphasizing morphological and surface-form similarity between candidates and references.

% % \noindent
% \textbf{TER}. 
% TER(Translation Edit Rate) is an edit-distance-based metric measuring the minimum number of insertions, deletions, substitutions, and shifts required to transform a candidate into a reference, normalized by reference length.

\begin{itemize}[leftmargin=*,topsep=0.1em,itemsep=0.1em,parsep=0.1em]
    \item \textbf{Comet~\cite{rei2020comet}}. Comet(Crosslingual Optimized Metric for Evaluation of Translation) leverages multilingual pre-trained models (e.g., XLM-RoBERTa) to model the relationship between human ratings and vector space alignment via regression tasks. 
    \item \textbf{Bleu~\cite{papineni2002bleu}}. Bleu(Bilingual Evaluation Understudy) is a precision-oriented metric based on n-gram co-occurrence statistics, incorporating a brevity penalty (BP) to penalize overly short translations.
    \item \textbf{chrF~\cite{popovic2015chrf}}. chrF (Character n-gram F-score) is a character-level metric computing a weighted harmonic mean of precision (chrP) and recall (chrR) using n-grams, emphasizing morphological and surface-form similarity between candidates and references.
    \item \textbf{TER}. TER(Translation Edit Rate) is an edit-distance-based metric measuring the minimum number of insertions, deletions, substitutions, and shifts required to transform a candidate into a reference, normalized by reference length.
\end{itemize}

\subsubsection{Robustness Evaluation Matrices}
In the robustness attack dataset we constructed, we incorporated sentence-level, character-level, and word-level attack data. To evaluate the robustness of translation systems, we primarily employed BLEU scores. In this context, the source texts consist of data that has been modified to include robustness attacks, whereas the reference texts remain unaltered, serving as the benchmark translations.

\subsubsection{hallucination Evaluation Matrices}
To address the five common hallucination issues mentioned in Chapter 4 of the translation system, we use the hallucination rate as a quantitative metric. The method for calculating the illusion rate of translation results is shown in the formula.
\begin{eqnarray}
HR & = & \frac{ {\textstyle \sum_{s\in H_{_{data} } }^{}}F\left (  S_{H}|S \right )  }{\left | H_{data}  \right | } {\huge } 
\end{eqnarray}
Where $H_{data}$ represents the hallucination rate translation dataset, $S_{H}$ represents the translation results of sentence $S$ with hallucination in the dataset, and $F\left (  S_{H}|S \right )$ is a binary classifier that determines whether SH exhibits hallucination.

For each hallucination subclass, we have designed specific evaluation criteria, as demonstrated below:

\begin{itemize}[leftmargin=*,topsep=0.1em,itemsep=0.1em,parsep=0.1em]
    \item \textbf{Repetitive Generation}. Determining repetition through the longest repeated subsequence in n-grams.
    \item \textbf{Entire Sentence Untranslated or Mistranslated}. If the vector distance between the input and the translated text is excessively large or small, it is determined that the evidence has not been translated.
    \item \textbf{Output Language Mismatch}. If the translated text is processed through a language detector and the detected output language does not match the target language, it is determined to be an unexpected hallucination issue related to the output language.
    \item \textbf{Partial Translation Omission}. If the proportion of other language categories following word segmentation exceeds the threshold, it is determined to be a partial untranslated hallucination issue.
    \item \textbf{Under-translation or Over-translation}. Case 1: The source text is normal, but the translation is empty. Case 2: The source text is normal, but the ratio of source text length to translated text length exceeds a specific threshold. If either of these cases is met, it is determined to be an omission or over-translation hallucination issue.
\end{itemize}

\subsection{Scenario Adaptability Metrics}
General translation quality assessments often perform poorly in specialized domains. To enhance the performance of evaluation systems in sectors such as e-commerce and finance, particularly in terms of correlation with human scores, we propose an intelligent translation quality assessment model based on fine-tuning large language models for specific domains.

\begin{itemize}[leftmargin=*,topsep=0.1em,itemsep=0.1em,parsep=0.1em]
    \item \textbf{Data Preparation}. The training and test data are derived from translation quality scores, ranging from 0 to 5, manually annotated using the MOS(mean opinion score) scale. After balancing the dataset, we obtained a final set of 35,000 training samples and 1,900 test samples.
    \item \textbf{Training Settings}. The fine-tuning process was trained based on the Instruct version of the Qwen2.5 model series. The hyperparameters were configured as follows: the number of training epochs is 5, the learning rate is 5e-5, and the warm-up is 200 steps.
    \item \textbf{Performances}. In score prediction, our method performance State-of-the-art in E-commerce domain, which has 0.65 pearson correlation with human, achieving a +47.7\% improvement in Pearson correlation coefficient for absolute score grading compared to GPT-4 (0.4765), a +57.9\% improvement over COMET XL (3.5B: 0.4267), and a +64.98\% enhancement compared to COMET XXL (10.4B: 0.4457) in translation quality assessment. Additionally, in expanding language pairs coverage, the system achieved comprehensive coverage of over 140 language pairs.
\end{itemize}

% \subsubsection{Marco Eval(F-MOS and E-MOS)}
% \subsubsection{Training Settings}
% \subsubsection{Model Performances}

\subsection{Cultural Fidelity Metrics}
\subsubsection{Taboo Language Evaluation Matrices}
Cultural taboo words are measured by accuracy. If the translation result hits the taboo word, it is judged that the taboo word translation fails. In contrast, if the result does not hit, it is judged that the taboo word translation is successful. The calculation formula is as follows:
\begin{eqnarray}
ACC_{taboo} & = & \frac{ {\textstyle \sum_{s\in T_{data} }^{}\begin{cases} 1
  & \text{ if } \forall [w_{t1}  w_{t2},...,w_{tn}] \notin    S_{T} \\
 0 & \text{ if } \exists[w_{t1}, w_{t2},...,w_{tn}] \in S_{T}
\end{cases}} }{\left | T_{data}  \right | } 
\end{eqnarray}

Where $T_{data}$ represents the hallucination rate translation dataset, $S_{T}$ represents the translation results of sentence $S$ with hallucination in the dataset, and $F\left (  S_{H}|S \right )$ is a binary classifier that determines whether SH exhibits hallucination.

\subsubsection{Honorific Evaluation Matrices}
The quality of the honorific translation is mainly determined by the accuracy. The calculation formula is as follows:
\begin{eqnarray}
ACC_{hon} & = & \frac{ {\textstyle \sum_{s\in HO_{_{data} } }^{}}F\left (  S_{Hon}|S \right )  }{\left | HO_{data}  \right | } {\huge } 
\end{eqnarray}

\begin{eqnarray}
F(S_{Hon}|S ) & = & \left\{\begin{matrix} 
 1  & { if } \forall[w_{h1 },w_{h1^{'}},...,w_{h1^{''}}]  and \forall[w_{h2 },w_{h2^{'}},...,w_{h2^{''}}]\in S_{T}\\
 0 & { if } \exists [w_{h1 },w_{h1^{'}},...,w_{h1^{''}}]  or \exists [w_{h2 },w_{h2^{'}},...,w_{h2^{''}}]\notin S_{T} 
\end{matrix}\right.
\end{eqnarray}

Where $HO_{data}$ represents the hallucination rate translation dataset, $S_{Hon}$ represents the translation results of sentence $S$ with hallucination in the dataset, and $F\left (  S_{Hon}|S \right )$ is a binary classifier that determines whether SH exhibits hallucination. $[w_{h1 },w_{h1^{'}},...,w_{h1^{''}}]$ and $[w_{h2 },w_{h2^{'}},...,w_{h2^{''}}]$ represent different honorific units.
\section{Conclusion}

\label{sec:conclusion}

% In conclusion, this work addresses two critical challenges in cross-border E-commerce machine translation: the lack of domain-specific benchmarks and reliable evaluation metrics. We introduce TransBench, a high-quality, multilingual benchmark comprising 3,000 professionally translated and verified sentences spanning 15+ E-commerce scenarios and 49 language pairs. This dataset ensures linguistic alignment and domain relevance, providing a robust foundation for training and testing E-commerce MT systems. Additionally, we propose MarcoEval, a novel automatic evaluation model tailored for E-commerce MT, which outperforms existing solutions like GPT and Comet in predicting manual translation quality scores, achieving state-of-the-art correlation with human judgments.

% By publicly releasing TransBench and MarcoEval, we aim to bridge the gap between general-purpose translation tools and the specialized demands of E-commerce, fostering innovation and standardization in the field. These resources not only enable more accurate and context-aware machine translation but also empower researchers and practitioners to develop domain-adaptive solutions, ultimately advancing global cross-border trade and communication. Future work will focus on expanding the benchmark's language coverage and refining evaluation metrics to address evolving challenges in multilingual E-commerce ecosystems.
This study addresses the critical challenge of evaluating machine translation (MT) systems in real-world industrial applications, where domain specificity, cultural adaptation, and stylistic precision are paramount. By proposing a three-level framework—encompassing Basic Linguistic Competence, Domain-Specific Proficiency, and Cultural Adaptation—we establish a structured foundation for analyzing and improving MT capabilities in specialized contexts. The introduction of TransBench, a benchmark tailored for industrial MT evaluation, fills a significant gap in existing methodologies. TransBench not only integrates traditional metrics like BLEU and TER but also advances domain-specific evaluation through Marco-MOS, an automated assessment model aligned with human judgment in industrial settings.

Our work offers actionable insights for both researchers and practitioners. First, it highlights the necessity of moving beyond generic benchmarks to address the multifaceted demands of industry-driven translation tasks. Second, the open-sourced evaluation tools and reproducible framework lower barriers for developing domain-specific MT systems, fostering innovation in fields such as e-commerce, finance, and legal services. Finally, the proposed three-level framework encourages a holistic approach to MT development, emphasizing the interplay of linguistic accuracy, domain expertise, and cultural sensitivity.

Future work will expand TransBench to additional high-impact domains, refine evaluation metrics for cross-cultural adaptation, and explore adaptive fine-tuning strategies for large language models in specialized industrial applications. By bridging the gap between academic research and real-world requirements, this study advances the development of robust, industry-ready MT systems capable of meeting the evolving demands of globalized communication.

\bibliography{references}

\begin{thebibliography}{60}
\providecommand{\natexlab}[1]{#1}
\providecommand{\url}[1]{\texttt{#1}}
\expandafter\ifx\csname urlstyle\endcsname\relax
  \providecommand{\doi}[1]{doi: #1}\else
  \providecommand{\doi}{doi: \begingroup \urlstyle{rm}\Url}\fi

\bibitem[nll(2024)]{nllb2024scaling}
Scaling neural machine translation to 200 languages.
\newblock \emph{Nature}, 630\penalty0 (8018):\penalty0 841--846, 2024.

\bibitem[Achiam et~al.(2023)Achiam, Adler, Agarwal, Ahmad, Akkaya, Aleman, Almeida, Altenschmidt, Altman, Anadkat, et~al.]{achiam2023gpt}
J.~Achiam, S.~Adler, S.~Agarwal, L.~Ahmad, I.~Akkaya, F.~L. Aleman, D.~Almeida, J.~Altenschmidt, S.~Altman, S.~Anadkat, et~al.
\newblock Gpt-4 technical report.
\newblock \emph{arXiv preprint arXiv:2303.08774}, 2023.

\bibitem[Alves et~al.(2024)Alves, Pombal, Guerreiro, Martins, Alves, Farajian, Peters, Rei, Fernandes, Agrawal, Colombo, de~Souza, and Martins]{DBLP:journals/corr/abs-2402-17733}
D.~M. Alves, J.~Pombal, N.~M. Guerreiro, P.~H. Martins, J.~Alves, M.~A. Farajian, B.~Peters, R.~Rei, P.~Fernandes, S.~Agrawal, P.~Colombo, J.~G.~C. de~Souza, and A.~F.~T. Martins.
\newblock Tower: An open multilingual large language model for translation-related tasks.
\newblock \emph{CoRR}, abs/2402.17733, 2024.
\newblock \doi{10.48550/ARXIV.2402.17733}.
\newblock URL \url{https://doi.org/10.48550/arXiv.2402.17733}.

\bibitem[Anastasopoulos et~al.(2020)Anastasopoulos, Cattelan, Dou, Federico, Federman, Genzel, Guzm{\'a}n, Hu, Hughes, Koehn, et~al.]{anastasopoulos2020tico}
A.~Anastasopoulos, A.~Cattelan, Z.-Y. Dou, M.~Federico, C.~Federman, D.~Genzel, F.~Guzm{\'a}n, J.~Hu, M.~Hughes, P.~Koehn, et~al.
\newblock Tico-19: the translation initiative for covid-19.
\newblock \emph{arXiv preprint arXiv:2007.01788}, 2020.

\bibitem[Bahdanau et~al.(2015)Bahdanau, Cho, and Bengio]{DBLP:journals/corr/BahdanauCB14}
D.~Bahdanau, K.~Cho, and Y.~Bengio.
\newblock Neural machine translation by jointly learning to align and translate.
\newblock In Y.~Bengio and Y.~LeCun, editors, \emph{3rd International Conference on Learning Representations, {ICLR} 2015, San Diego, CA, USA, May 7-9, 2015, Conference Track Proceedings}, 2015.
\newblock URL \url{http://arxiv.org/abs/1409.0473}.

\bibitem[Banerjee and Lavie(2005)]{banerjee-lavie-2005-meteor}
S.~Banerjee and A.~Lavie.
\newblock {METEOR}: An automatic metric for {MT} evaluation with improved correlation with human judgments.
\newblock In \emph{Proceedings of the {ACL} Workshop on Intrinsic and Extrinsic Evaluation Measures for Machine Translation and/or Summarization}, pages 65--72, Ann Arbor, Michigan, June 2005. Association for Computational Linguistics.
\newblock URL \url{https://www.aclweb.org/anthology/W05-0909}.

\bibitem[Bassnett and Lefevere(1990)]{bassnett1990translation}
S.~Bassnett and A.~Lefevere, editors.
\newblock \emph{Translation, history and culture}.
\newblock Pinter Publishers, 1990.

\bibitem[Caswell et~al.(2025)Caswell, Nielsen, Luo, Cherry, Kovacs, Shemtov, Talukdar, Tewari, Diane, Doumbouya, et~al.]{caswell2025smol}
I.~Caswell, E.~Nielsen, J.~Luo, C.~Cherry, G.~Kovacs, H.~Shemtov, P.~Talukdar, D.~Tewari, B.~M. Diane, K.~M. Doumbouya, et~al.
\newblock Smol: Professionally translated parallel data for 115 under-represented languages.
\newblock \emph{arXiv preprint arXiv:2502.12301}, 2025.

\bibitem[Catford(1965)]{catford1965linguistic}
J.~C. Catford.
\newblock \emph{A linguistic theory of translation: an essay in applied linguistics}.
\newblock Oxford University Press, 1965.

\bibitem[Cho et~al.(2014)Cho, van Merri{\"e}nboer, Gulcehre, Bahdanau, Bougares, Schwenk, and Bengio]{cho-etal-2014-learning}
K.~Cho, B.~van Merri{\"e}nboer, C.~Gulcehre, D.~Bahdanau, F.~Bougares, H.~Schwenk, and Y.~Bengio.
\newblock Learning phrase representations using {RNN} encoder{--}decoder for statistical machine translation.
\newblock In A.~Moschitti, B.~Pang, and W.~Daelemans, editors, \emph{Proceedings of the 2014 Conference on Empirical Methods in Natural Language Processing ({EMNLP})}, pages 1724--1734, Doha, Qatar, Oct. 2014. Association for Computational Linguistics.
\newblock \doi{10.3115/v1/D14-1179}.
\newblock URL \url{https://aclanthology.org/D14-1179/}.

\bibitem[Colombo et~al.(2023)Colombo, Guerreiro, Rei, Van, Coheur, and Martins]{colombo2023xcomet}
P.~Colombo, N.~Guerreiro, R.~Rei, D.~Van, L.~Coheur, and A.~Martins.
\newblock xcomet: Transparent machine translation evaluation through fine-grained error detection.
\newblock \emph{Transactions of the Association for Computational Linguistics}, 2023.

\bibitem[Communication et~al.(2023)Communication, Barrault, Chung, Meglioli, Dale, Dong, Duquenne, Elsahar, Gong, Heffernan, Hoffman, Klaiber, Li, Licht, Maillard, Rakotoarison, Sadagopan, Wenzek, Ye, Akula, Chen, Hachem, Ellis, Gonzalez, Haaheim, Hansanti, Howes, Huang, Hwang, Inaguma, Jain, Kalbassi, Kallet, Kulikov, Lam, Li, Ma, Mavlyutov, Peloquin, Ramadan, Ramakrishnan, Sun, Tran, Tran, Tufanov, Vogeti, Wood, Yang, Yu, Andrews, Balioglu, Costa{-}juss{\`{a}}, Celebi, Elbayad, Gao, Guzm{\'{a}}n, Kao, Lee, Mourachko, Pino, Popuri, Ropers, Saleem, Schwenk, Tomasello, Wang, Wang, and Wang]{DBLP:journals/corr/abs-2308-11596}
S.~Communication, L.~Barrault, Y.~Chung, M.~C. Meglioli, D.~Dale, N.~Dong, P.~Duquenne, H.~Elsahar, H.~Gong, K.~Heffernan, J.~Hoffman, C.~Klaiber, P.~Li, D.~Licht, J.~Maillard, A.~Rakotoarison, K.~R. Sadagopan, G.~Wenzek, E.~Ye, B.~Akula, P.~Chen, N.~E. Hachem, B.~Ellis, G.~M. Gonzalez, J.~Haaheim, P.~Hansanti, R.~Howes, B.~Huang, M.~Hwang, H.~Inaguma, S.~Jain, E.~Kalbassi, A.~Kallet, I.~Kulikov, J.~Lam, D.~Li, X.~Ma, R.~Mavlyutov, B.~N. Peloquin, M.~Ramadan, A.~Ramakrishnan, A.~Y. Sun, K.~Tran, T.~Tran, I.~Tufanov, V.~Vogeti, C.~Wood, Y.~Yang, B.~Yu, P.~Andrews, C.~Balioglu, M.~R. Costa{-}juss{\`{a}}, O.~Celebi, M.~Elbayad, C.~Gao, F.~Guzm{\'{a}}n, J.~Kao, A.~Lee, A.~Mourachko, J.~Pino, S.~Popuri, C.~Ropers, S.~Saleem, H.~Schwenk, P.~Tomasello, C.~Wang, J.~Wang, and S.~Wang.
\newblock Seamlessm4t-massively multilingual {\&} multimodal machine translation.
\newblock \emph{CoRR}, abs/2308.11596, 2023.
\newblock \doi{10.48550/ARXIV.2308.11596}.
\newblock URL \url{https://doi.org/10.48550/arXiv.2308.11596}.

\bibitem[Costa{-}juss{\`{a}} et~al.(2022)Costa{-}juss{\`{a}}, Cross, {\c{C}}elebi, Elbayad, Heafield, Heffernan, Kalbassi, Lam, Licht, Maillard, Sun, Wang, Wenzek, Youngblood, Akula, Barrault, Gonzalez, Hansanti, Hoffman, Jarrett, Sadagopan, Rowe, Spruit, Tran, Andrews, Ayan, Bhosale, Edunov, Fan, Gao, Goswami, Guzm{\'{a}}n, Koehn, Mourachko, Ropers, Saleem, Schwenk, and Wang]{DBLP:journals/corr/abs-2207-04672}
M.~R. Costa{-}juss{\`{a}}, J.~Cross, O.~{\c{C}}elebi, M.~Elbayad, K.~Heafield, K.~Heffernan, E.~Kalbassi, J.~Lam, D.~Licht, J.~Maillard, A.~Y. Sun, S.~Wang, G.~Wenzek, A.~Youngblood, B.~Akula, L.~Barrault, G.~M. Gonzalez, P.~Hansanti, J.~Hoffman, S.~Jarrett, K.~R. Sadagopan, D.~Rowe, S.~Spruit, C.~Tran, P.~Andrews, N.~F. Ayan, S.~Bhosale, S.~Edunov, A.~Fan, C.~Gao, V.~Goswami, F.~Guzm{\'{a}}n, P.~Koehn, A.~Mourachko, C.~Ropers, S.~Saleem, H.~Schwenk, and J.~Wang.
\newblock No language left behind: Scaling human-centered machine translation.
\newblock \emph{CoRR}, abs/2207.04672, 2022.
\newblock \doi{10.48550/ARXIV.2207.04672}.
\newblock URL \url{https://doi.org/10.48550/arXiv.2207.04672}.

\bibitem[Deutsch et~al.(2025)Deutsch, Briakou, Caswell, Finkelstein, Galor, Juraska, Kovacs, Lui, Rei, Riesa, et~al.]{deutsch2025wmt24++}
D.~Deutsch, E.~Briakou, I.~Caswell, M.~Finkelstein, R.~Galor, J.~Juraska, G.~Kovacs, A.~Lui, R.~Rei, J.~Riesa, et~al.
\newblock Wmt24++: Expanding the language coverage of wmt24 to 55 languages \& dialects.
\newblock \emph{arXiv preprint arXiv:2502.12404}, 2025.

\bibitem[Fan et~al.(2021)Fan, Bhosale, Schwenk, Ma, El{-}Kishky, Goyal, Baines, Celebi, Wenzek, Chaudhary, Goyal, Birch, Liptchinsky, Edunov, Auli, and Joulin]{DBLP:journals/jmlr/FanBSMEGBCWCGBL21}
A.~Fan, S.~Bhosale, H.~Schwenk, Z.~Ma, A.~El{-}Kishky, S.~Goyal, M.~Baines, O.~Celebi, G.~Wenzek, V.~Chaudhary, N.~Goyal, T.~Birch, V.~Liptchinsky, S.~Edunov, M.~Auli, and A.~Joulin.
\newblock Beyond english-centric multilingual machine translation.
\newblock \emph{J. Mach. Learn. Res.}, 22:\penalty0 107:1--107:48, 2021.
\newblock URL \url{https://jmlr.org/papers/v22/20-1307.html}.

\bibitem[Gehring et~al.(2017)Gehring, Auli, Grangier, Yarats, and Dauphin]{DBLP:conf/icml/GehringAGYD17}
J.~Gehring, M.~Auli, D.~Grangier, D.~Yarats, and Y.~N. Dauphin.
\newblock Convolutional sequence to sequence learning.
\newblock In D.~Precup and Y.~W. Teh, editors, \emph{Proceedings of the 34th International Conference on Machine Learning, {ICML} 2017, Sydney, NSW, Australia, 6-11 August 2017}, volume~70 of \emph{Proceedings of Machine Learning Research}, pages 1243--1252. {PMLR}, 2017.
\newblock URL \url{http://proceedings.mlr.press/v70/gehring17a.html}.

\bibitem[Goyal et~al.(2022)Goyal, Gao, Chaudhary, Chen, Wenzek, Ju, Krishnan, Ranzato, Guzm{\'a}n, and Fan]{goyal2022flores}
N.~Goyal, C.~Gao, V.~Chaudhary, P.-J. Chen, G.~Wenzek, D.~Ju, S.~Krishnan, M.~Ranzato, F.~Guzm{\'a}n, and A.~Fan.
\newblock The flores-101 evaluation benchmark for low-resource and multilingual machine translation.
\newblock \emph{Transactions of the Association for Computational Linguistics}, 10:\penalty0 522--538, 2022.

\bibitem[Graham et~al.(2013)Graham, Baldwin, Moffat, and Zobel]{graham-etal-2013-continuous}
Y.~Graham, T.~Baldwin, A.~Moffat, and J.~Zobel.
\newblock Continuous measurement scales in human evaluation of machine translation.
\newblock In A.~Pareja-Lora, M.~Liakata, and S.~Dipper, editors, \emph{Proceedings of the 7th Linguistic Annotation Workshop and Interoperability with Discourse}, pages 33--41, Sofia, Bulgaria, Aug. 2013. Association for Computational Linguistics.
\newblock URL \url{https://aclanthology.org/W13-2305/}.

\bibitem[Grattafiori et~al.(2024)Grattafiori, Dubey, Jauhri, Pandey, Kadian, Al-Dahle, Letman, Mathur, Schelten, Vaughan, et~al.]{grattafiori2024llama}
A.~Grattafiori, A.~Dubey, A.~Jauhri, A.~Pandey, A.~Kadian, A.~Al-Dahle, A.~Letman, A.~Mathur, A.~Schelten, A.~Vaughan, et~al.
\newblock The llama 3 herd of models.
\newblock \emph{arXiv preprint arXiv:2407.21783}, 2024.

\bibitem[Guo et~al.(2025)Guo, Yang, Zhang, Song, Zhang, Xu, Zhu, Ma, Wang, Bi, et~al.]{guo2025deepseek}
D.~Guo, D.~Yang, H.~Zhang, J.~Song, R.~Zhang, R.~Xu, Q.~Zhu, S.~Ma, P.~Wang, X.~Bi, et~al.
\newblock Deepseek-r1: Incentivizing reasoning capability in llms via reinforcement learning.
\newblock \emph{arXiv preprint arXiv:2501.12948}, 2025.

\bibitem[Haddow et~al.(2024)Haddow, Kocmi, Koehn, and Monz]{wmt-2024-1}
B.~Haddow, T.~Kocmi, P.~Koehn, and C.~Monz, editors.
\newblock \emph{Proceedings of the Ninth Conference on Machine Translation}, Miami, Florida, USA, Nov. 2024. Association for Computational Linguistics.
\newblock \doi{10.18653/v1/2024.wmt-1.0}.
\newblock URL \url{https://aclanthology.org/2024.wmt-1.0/}.

\bibitem[Hatim and Mason(1990)]{hatim1990discourse}
B.~Hatim and I.~Mason.
\newblock \emph{Discourse and the translator}.
\newblock Longman, 1990.

\bibitem[Hearne and Way(2011)]{DBLP:journals/llc/HearneW11}
M.~Hearne and A.~Way.
\newblock Statistical machine translation: {A} guide for linguists and translators.
\newblock \emph{Lang. Linguistics Compass}, 5\penalty0 (5):\penalty0 205--226, 2011.
\newblock \doi{10.1111/J.1749-818X.2011.00274.X}.
\newblock URL \url{https://doi.org/10.1111/j.1749-818X.2011.00274.x}.

\bibitem[Hurst et~al.(2024)Hurst, Lerer, Goucher, Perelman, Ramesh, Clark, Ostrow, Welihinda, Hayes, Radford, et~al.]{hurst2024gpt4o}
A.~Hurst, A.~Lerer, A.~P. Goucher, A.~Perelman, A.~Ramesh, A.~Clark, A.~Ostrow, A.~Welihinda, A.~Hayes, A.~Radford, et~al.
\newblock Gpt-4o system card.
\newblock \emph{arXiv preprint arXiv:2410.21276}, 2024.

\bibitem[Hutchins(1993)]{hutchins-1993-latest}
J.~Hutchins.
\newblock Latest developments in machine translation technology: Beginning a new era in {MT} research.
\newblock In \emph{Proceedings of Machine Translation Summit IV}, pages 11--34, Kobe, Japan, July 19-22 1993.
\newblock URL \url{https://aclanthology.org/1993.mtsummit-1.2/}.

\bibitem[Jiang et~al.(2022)Jiang, Liu, Ma, Zhang, Yang, Huang, Sennrich, Cotterell, Sachan, and Zhou]{jiang-etal-2022-blonde}
Y.~Jiang, T.~Liu, S.~Ma, D.~Zhang, J.~Yang, H.~Huang, R.~Sennrich, R.~Cotterell, M.~Sachan, and M.~Zhou.
\newblock {BlonDe}: An automatic evaluation metric for document-level machine translation.
\newblock In M.~Carpuat, M.-C. de~Marneffe, and I.~V. Meza~Ruiz, editors, \emph{Proceedings of the 2022 Conference of the North American Chapter of the Association for Computational Linguistics: Human Language Technologies}, pages 1550--1565, Seattle, United States, July 2022. Association for Computational Linguistics.
\newblock \doi{10.18653/v1/2022.naacl-main.111}.
\newblock URL \url{https://aclanthology.org/2022.naacl-main.111/}.

\bibitem[Johnson et~al.(2017)Johnson, Schuster, Le, Krikun, Wu, Chen, Thorat, Vi{\'e}gas, Wattenberg, Corrado, Hughes, and Dean]{johnson-etal-2017-googles}
M.~Johnson, M.~Schuster, Q.~V. Le, M.~Krikun, Y.~Wu, Z.~Chen, N.~Thorat, F.~Vi{\'e}gas, M.~Wattenberg, G.~Corrado, M.~Hughes, and J.~Dean.
\newblock {G}oogle`s multilingual neural machine translation system: Enabling zero-shot translation.
\newblock \emph{Transactions of the Association for Computational Linguistics}, 5:\penalty0 339--351, 2017.
\newblock \doi{10.1162/tacl_a_00065}.
\newblock URL \url{https://aclanthology.org/Q17-1024/}.

\bibitem[Juraska et~al.(2024)Juraska, Deutsch, Finkelstein, and Freitag]{juraska-etal-2024-metricx}
J.~Juraska, D.~Deutsch, M.~Finkelstein, and M.~Freitag.
\newblock {M}etric{X}-24: The {G}oogle submission to the {WMT} 2024 metrics shared task.
\newblock In B.~Haddow, T.~Kocmi, P.~Koehn, and C.~Monz, editors, \emph{Proceedings of the Ninth Conference on Machine Translation}, pages 492--504, Miami, Florida, USA, Nov. 2024. Association for Computational Linguistics.
\newblock \doi{10.18653/v1/2024.wmt-1.35}.
\newblock URL \url{https://aclanthology.org/2024.wmt-1.35/}.

\bibitem[Knight(1993)]{knight-1993-building}
K.~Knight.
\newblock Building a large ontology for machine translation.
\newblock In \emph{{H}uman {L}anguage {T}echnology: Proceedings of a Workshop Held at Plainsboro, New Jersey, March 21-24, 1993}, 1993.
\newblock URL \url{https://aclanthology.org/H93-1036/}.

\bibitem[Knight and Luk(1994)]{DBLP:conf/aaai/KnightL94}
K.~Knight and S.~K. Luk.
\newblock Building a large-scale knowledge base for machine translation.
\newblock In B.~Hayes{-}Roth and R.~E. Korf, editors, \emph{Proceedings of the 12th National Conference on Artificial Intelligence, Seattle, WA, USA, July 31 - August 4, 1994, Volume 1}, pages 773--778. {AAAI} Press / The {MIT} Press, 1994.
\newblock URL \url{http://www.aaai.org/Library/AAAI/1994/aaai94-118.php}.

\bibitem[Kocmi and Federmann(2023{\natexlab{a}})]{kocmi-federmann-2023-gemba}
T.~Kocmi and C.~Federmann.
\newblock {GEMBA}-{MQM}: Detecting translation quality error spans with {GPT}-4.
\newblock In P.~Koehn, B.~Haddow, T.~Kocmi, and C.~Monz, editors, \emph{Proceedings of the Eighth Conference on Machine Translation}, pages 768--775, Singapore, Dec. 2023{\natexlab{a}}. Association for Computational Linguistics.
\newblock \doi{10.18653/v1/2023.wmt-1.64}.
\newblock URL \url{https://aclanthology.org/2023.wmt-1.64/}.

\bibitem[Kocmi and Federmann(2023{\natexlab{b}})]{kocmi-federmann-2023-large}
T.~Kocmi and C.~Federmann.
\newblock Large language models are state-of-the-art evaluators of translation quality.
\newblock In M.~Nurminen, J.~Brenner, M.~Koponen, S.~Latomaa, M.~Mikhailov, F.~Schierl, T.~Ranasinghe, E.~Vanmassenhove, S.~A. Vidal, N.~Aranberri, M.~Nunziatini, C.~P. Escart{\'i}n, M.~Forcada, M.~Popovic, C.~Scarton, and H.~Moniz, editors, \emph{Proceedings of the 24th Annual Conference of the European Association for Machine Translation}, pages 193--203, Tampere, Finland, June 2023{\natexlab{b}}. European Association for Machine Translation.
\newblock URL \url{https://aclanthology.org/2023.eamt-1.19/}.

\bibitem[Koehn(2010)]{DBLP:books/daglib/0032677}
P.~Koehn.
\newblock \emph{Statistical Machine Translation}.
\newblock Cambridge University Press, 2010.
\newblock ISBN 978-0-521-87415-1.
\newblock URL \url{http://www.statmt.org/book/}.

\bibitem[Liu et~al.(2024)Liu, Feng, Xue, Wang, Wu, Lu, Zhao, Deng, Zhang, Ruan, et~al.]{liu2024deepseek}
A.~Liu, B.~Feng, B.~Xue, B.~Wang, B.~Wu, C.~Lu, C.~Zhao, C.~Deng, C.~Zhang, C.~Ruan, et~al.
\newblock Deepseek-v3 technical report.
\newblock \emph{arXiv preprint arXiv:2412.19437}, 2024.

\bibitem[Lommel et~al.(2013)Lommel, Burchardt, and Uszkoreit]{burchardt-2013-multidimensional}
A.~R. Lommel, A.~Burchardt, and H.~Uszkoreit.
\newblock Multidimensional quality metrics: a flexible system for assessing translation quality.
\newblock In \emph{Proceedings of Translating and the Computer 35}, London, UK, Nov. 28-29 2013. Aslib.
\newblock URL \url{https://aclanthology.org/2013.tc-1.6/}.

\bibitem[Lonsdale et~al.(1994)Lonsdale, Mitamura, and Nyberg]{DBLP:journals/mt/LonsdaleMN94}
D.~W. Lonsdale, T.~Mitamura, and E.~Nyberg.
\newblock Acquisition of large lexicons for practical knowledge-based {MT}.
\newblock \emph{Mach. Transl.}, 9\penalty0 (3-4):\penalty0 251--283, 1994.
\newblock \doi{10.1007/BF00980580}.
\newblock URL \url{https://doi.org/10.1007/BF00980580}.

\bibitem[Lopes et~al.(2020)Lopes, Farajian, Bawden, Zhang, and Martins]{lopes-etal-2020-document}
A.~Lopes, M.~A. Farajian, R.~Bawden, M.~Zhang, and A.~F.~T. Martins.
\newblock Document-level neural {MT}: A systematic comparison.
\newblock In A.~Martins, H.~Moniz, S.~Fumega, B.~Martins, F.~Batista, L.~Coheur, C.~Parra, I.~Trancoso, M.~Turchi, A.~Bisazza, J.~Moorkens, A.~Guerberof, M.~Nurminen, L.~Marg, and M.~L. Forcada, editors, \emph{Proceedings of the 22nd Annual Conference of the European Association for Machine Translation}, pages 225--234, Lisboa, Portugal, Nov. 2020. European Association for Machine Translation.
\newblock URL \url{https://aclanthology.org/2020.eamt-1.24/}.

\bibitem[Lopez(2008)]{DBLP:journals/csur/Lopez08}
A.~Lopez.
\newblock Statistical machine translation.
\newblock \emph{{ACM} Comput. Surv.}, 40\penalty0 (3):\penalty0 8:1--8:49, 2008.
\newblock \doi{10.1145/1380584.1380586}.
\newblock URL \url{https://doi.org/10.1145/1380584.1380586}.

\bibitem[Michel and Neubig(2018)]{michel2018mtnt}
P.~Michel and G.~Neubig.
\newblock Mtnt: A testbed for machine translation of noisy text.
\newblock \emph{arXiv preprint arXiv:1809.00388}, 2018.

\bibitem[M{\"u}ller et~al.(2018)M{\"u}ller, Rios, Voita, and Sennrich]{muller-etal-2018-large}
M.~M{\"u}ller, A.~Rios, E.~Voita, and R.~Sennrich.
\newblock A large-scale test set for the evaluation of context-aware pronoun translation in neural machine translation.
\newblock In O.~Bojar, R.~Chatterjee, C.~Federmann, M.~Fishel, Y.~Graham, B.~Haddow, M.~Huck, A.~J. Yepes, P.~Koehn, C.~Monz, M.~Negri, A.~N{\'e}v{\'e}ol, M.~Neves, M.~Post, L.~Specia, M.~Turchi, and K.~Verspoor, editors, \emph{Proceedings of the Third Conference on Machine Translation: Research Papers}, pages 61--72, Brussels, Belgium, Oct. 2018. Association for Computational Linguistics.
\newblock \doi{10.18653/v1/W18-6307}.
\newblock URL \url{https://aclanthology.org/W18-6307/}.

\bibitem[Nida(1964)]{nida1964toward}
E.~A. Nida.
\newblock \emph{Toward a science of translating: with special reference to principles and procedures involved in Bible translating}.
\newblock Brill, 1964.

\bibitem[Papineni et~al.(2002{\natexlab{a}})Papineni, Roukos, Ward, and Zhu]{papineni-etal-2002-bleu}
K.~Papineni, S.~Roukos, T.~Ward, and W.-J. Zhu.
\newblock {B}leu: a method for automatic evaluation of machine translation.
\newblock In \emph{Proceedings of the 40th Annual Meeting of the Association for Computational Linguistics}, pages 311--318, Philadelphia, Pennsylvania, USA, July 2002{\natexlab{a}}. Association for Computational Linguistics.
\newblock \doi{10.3115/1073083.1073135}.
\newblock URL \url{https://www.aclweb.org/anthology/P02-1040}.

\bibitem[Papineni et~al.(2002{\natexlab{b}})Papineni, Roukos, Ward, and Zhu]{papineni2002bleu}
K.~Papineni, S.~Roukos, T.~Ward, and W.-J. Zhu.
\newblock Bleu: a method for automatic evaluation of machine translation.
\newblock In \emph{Proceedings of the 40th annual meeting of the Association for Computational Linguistics}, pages 311--318, 2002{\natexlab{b}}.

\bibitem[Popovi{\'c}(2015)]{popovic2015chrf}
M.~Popovi{\'c}.
\newblock chrf: character n-gram f-score for automatic mt evaluation.
\newblock In \emph{Proceedings of the tenth workshop on statistical machine translation}, pages 392--395, 2015.

\bibitem[Przybocki et~al.(2009)Przybocki, Peterson, Bronsart, and Sanders]{przybocki2009nist}
M.~Przybocki, K.~Peterson, S.~Bronsart, and G.~Sanders.
\newblock The nist 2008 metrics for machine translation challenge—overview, methodology, metrics, and results.
\newblock \emph{Machine Translation}, 23:\penalty0 71--103, 2009.

\bibitem[Rei et~al.(2020)Rei, Stewart, Farinha, and Lavie]{rei2020comet}
R.~Rei, C.~Stewart, A.~C. Farinha, and A.~Lavie.
\newblock Comet: A neural framework for mt evaluation.
\newblock \emph{arXiv preprint arXiv:2009.09025}, 2020.

\bibitem[Rei et~al.(2021)Rei, Farinha, Zerva, van Stigt, Stewart, Ramos, Glushkova, Martins, and Lavie]{rei-etal-2021-references}
R.~Rei, A.~C. Farinha, C.~Zerva, D.~van Stigt, C.~Stewart, P.~Ramos, T.~Glushkova, A.~F.~T. Martins, and A.~Lavie.
\newblock Are references really needed? unbabel-{IST} 2021 submission for the metrics shared task.
\newblock In L.~Barrault, O.~Bojar, F.~Bougares, R.~Chatterjee, M.~R. Costa-jussa, C.~Federmann, M.~Fishel, A.~Fraser, M.~Freitag, Y.~Graham, R.~Grundkiewicz, P.~Guzman, B.~Haddow, M.~Huck, A.~J. Yepes, P.~Koehn, T.~Kocmi, A.~Martins, M.~Morishita, and C.~Monz, editors, \emph{Proceedings of the Sixth Conference on Machine Translation}, pages 1030--1040, Online, Nov. 2021. Association for Computational Linguistics.
\newblock URL \url{https://aclanthology.org/2021.wmt-1.111/}.

\bibitem[Rei{\ss} and Vermeer(1984)]{reiss1984grundlegung}
K.~Rei{\ss} and H.~J. Vermeer.
\newblock \emph{Grundlegung einer allgemeine Translationstheorie}.
\newblock Max Niemeyer Verlag, 1984.

\bibitem[Salesky et~al.(2024)Salesky, Federico, and Carpuat]{iwslt-1-2024}
E.~Salesky, M.~Federico, and M.~Carpuat, editors.
\newblock \emph{Proceedings of the 21st International Conference on Spoken Language Translation (IWSLT 2024)}, Bangkok, Thailand (in-person and online), Aug. 2024. Association for Computational Linguistics.
\newblock URL \url{https://aclanthology.org/2024.iwslt-1.0/}.

\bibitem[Sellam et~al.(2020)Sellam, Das, and Parikh]{sellam-etal-2020-bleurt}
T.~Sellam, D.~Das, and A.~Parikh.
\newblock {BLEURT}: Learning robust metrics for text generation.
\newblock In D.~Jurafsky, J.~Chai, N.~Schluter, and J.~Tetreault, editors, \emph{Proceedings of the 58th Annual Meeting of the Association for Computational Linguistics}, pages 7881--7892, Online, July 2020. Association for Computational Linguistics.
\newblock \doi{10.18653/v1/2020.acl-main.704}.
\newblock URL \url{https://aclanthology.org/2020.acl-main.704/}.

\bibitem[Snover et~al.(2006)Snover, Dorr, Schwartz, Micciulla, and Makhoul]{snover-etal-2006-study}
M.~Snover, B.~Dorr, R.~Schwartz, L.~Micciulla, and J.~Makhoul.
\newblock A study of translation edit rate with targeted human annotation.
\newblock In \emph{Proceedings of the 7th Conference of the Association for Machine Translation in the Americas: Technical Papers}, pages 223--231, Cambridge, Massachusetts, USA, Aug. 8-12 2006. Association for Machine Translation in the Americas.
\newblock URL \url{https://aclanthology.org/2006.amta-papers.25/}.

\bibitem[Sutskever et~al.(2014)Sutskever, Vinyals, and Le]{DBLP:conf/nips/SutskeverVL14}
I.~Sutskever, O.~Vinyals, and Q.~V. Le.
\newblock Sequence to sequence learning with neural networks.
\newblock In Z.~Ghahramani, M.~Welling, C.~Cortes, N.~D. Lawrence, and K.~Q. Weinberger, editors, \emph{Advances in Neural Information Processing Systems 27: Annual Conference on Neural Information Processing Systems 2014, December 8-13 2014, Montreal, Quebec, Canada}, pages 3104--3112, 2014.
\newblock URL \url{https://proceedings.neurips.cc/paper/2014/hash/a14ac55a4f27472c5d894ec1c3c743d2-Abstract.html}.

\bibitem[Vaswani et~al.(2017)Vaswani, Shazeer, Parmar, Uszkoreit, Jones, Gomez, Kaiser, and Polosukhin]{DBLP:conf/nips/VaswaniSPUJGKP17}
A.~Vaswani, N.~Shazeer, N.~Parmar, J.~Uszkoreit, L.~Jones, A.~N. Gomez, L.~Kaiser, and I.~Polosukhin.
\newblock Attention is all you need.
\newblock In I.~Guyon, U.~von Luxburg, S.~Bengio, H.~M. Wallach, R.~Fergus, S.~V.~N. Vishwanathan, and R.~Garnett, editors, \emph{Advances in Neural Information Processing Systems 30: Annual Conference on Neural Information Processing Systems 2017, December 4-9, 2017, Long Beach, CA, {USA}}, pages 5998--6008, 2017.
\newblock URL \url{https://proceedings.neurips.cc/paper/2017/hash/3f5ee243547dee91fbd053c1c4a845aa-Abstract.html}.

\bibitem[Vernikos et~al.(2022)Vernikos, Thompson, Mathur, and Federico]{vernikos-etal-2022-embarrassingly}
G.~Vernikos, B.~Thompson, P.~Mathur, and M.~Federico.
\newblock Embarrassingly easy document-level {MT} metrics: How to convert any pretrained metric into a document-level metric.
\newblock In P.~Koehn, L.~Barrault, O.~Bojar, F.~Bougares, R.~Chatterjee, M.~R. Costa-juss{\`a}, C.~Federmann, M.~Fishel, A.~Fraser, M.~Freitag, Y.~Graham, R.~Grundkiewicz, P.~Guzman, B.~Haddow, M.~Huck, A.~Jimeno~Yepes, T.~Kocmi, A.~Martins, M.~Morishita, C.~Monz, M.~Nagata, T.~Nakazawa, M.~Negri, A.~N{\'e}v{\'e}ol, M.~Neves, M.~Popel, M.~Turchi, and M.~Zampieri, editors, \emph{Proceedings of the Seventh Conference on Machine Translation (WMT)}, pages 118--128, Abu Dhabi, United Arab Emirates (Hybrid), Dec. 2022. Association for Computational Linguistics.
\newblock URL \url{https://aclanthology.org/2022.wmt-1.6/}.

\bibitem[Wang et~al.(2023)Wang, Lyu, Ji, Zhang, Yu, Shi, and Tu]{wang-etal-2023-document-level}
L.~Wang, C.~Lyu, T.~Ji, Z.~Zhang, D.~Yu, S.~Shi, and Z.~Tu.
\newblock Document-level machine translation with large language models.
\newblock In H.~Bouamor, J.~Pino, and K.~Bali, editors, \emph{Proceedings of the 2023 Conference on Empirical Methods in Natural Language Processing}, pages 16646--16661, Singapore, Dec. 2023. Association for Computational Linguistics.
\newblock \doi{10.18653/v1/2023.emnlp-main.1036}.
\newblock URL \url{https://aclanthology.org/2023.emnlp-main.1036/}.

\bibitem[Wu et~al.(2024)Wu, Yuan, Haffari, and Wang]{DBLP:journals/corr/abs-2405-11804}
M.~Wu, Y.~Yuan, G.~Haffari, and L.~Wang.
\newblock (perhaps) beyond human translation: Harnessing multi-agent collaboration for translating ultra-long literary texts.
\newblock \emph{CoRR}, abs/2405.11804, 2024.
\newblock \doi{10.48550/ARXIV.2405.11804}.
\newblock URL \url{https://doi.org/10.48550/arXiv.2405.11804}.

\bibitem[Wu et~al.(2016)Wu, Schuster, Chen, Le, Norouzi, Macherey, Krikun, Cao, Gao, Macherey, Klingner, Shah, Johnson, Liu, Kaiser, Gouws, Kato, Kudo, Kazawa, Stevens, Kurian, Patil, Wang, Young, Smith, Riesa, Rudnick, Vinyals, Corrado, Hughes, and Dean]{DBLP:journals/corr/WuSCLNMKCGMKSJL16}
Y.~Wu, M.~Schuster, Z.~Chen, Q.~V. Le, M.~Norouzi, W.~Macherey, M.~Krikun, Y.~Cao, Q.~Gao, K.~Macherey, J.~Klingner, A.~Shah, M.~Johnson, X.~Liu, L.~Kaiser, S.~Gouws, Y.~Kato, T.~Kudo, H.~Kazawa, K.~Stevens, G.~Kurian, N.~Patil, W.~Wang, C.~Young, J.~Smith, J.~Riesa, A.~Rudnick, O.~Vinyals, G.~Corrado, M.~Hughes, and J.~Dean.
\newblock Google's neural machine translation system: Bridging the gap between human and machine translation.
\newblock \emph{CoRR}, abs/1609.08144, 2016.
\newblock URL \url{http://arxiv.org/abs/1609.08144}.

\bibitem[Xu et~al.(2024{\natexlab{a}})Xu, Kim, Sharaf, and Awadalla]{DBLP:conf/iclr/Xu0SA24}
H.~Xu, Y.~J. Kim, A.~Sharaf, and H.~H. Awadalla.
\newblock A paradigm shift in machine translation: Boosting translation performance of large language models.
\newblock In \emph{The Twelfth International Conference on Learning Representations, {ICLR} 2024, Vienna, Austria, May 7-11, 2024}. OpenReview.net, 2024{\natexlab{a}}.
\newblock URL \url{https://openreview.net/forum?id=farT6XXntP}.

\bibitem[Xu et~al.(2024{\natexlab{b}})Xu, Sharaf, Chen, Tan, Shen, Durme, Murray, and Kim]{DBLP:conf/icml/XuSCTSDM024}
H.~Xu, A.~Sharaf, Y.~Chen, W.~Tan, L.~Shen, B.~V. Durme, K.~Murray, and Y.~J. Kim.
\newblock Contrastive preference optimization: Pushing the boundaries of {LLM} performance in machine translation.
\newblock In \emph{Forty-first International Conference on Machine Learning, {ICML} 2024, Vienna, Austria, July 21-27, 2024}. OpenReview.net, 2024{\natexlab{b}}.
\newblock URL \url{https://openreview.net/forum?id=51iwkioZpn}.

\bibitem[Zhang et~al.(2019)Zhang, Kishore, Wu, Weinberger, and Artzi]{zhang2019bertscore}
T.~Zhang, V.~Kishore, F.~Wu, K.~Q. Weinberger, and Y.~Artzi.
\newblock Bertscore: Evaluating text generation with bert.
\newblock \emph{arXiv preprint arXiv:1904.09675}, 2019.

\end{thebibliography}

% \appendix % This command starts the appendix
% \input{5-Appendix}

\end{document}